\documentclass{article}

\usepackage{microtype}
\usepackage{graphicx}
\usepackage{subfigure}
\usepackage{booktabs} 

\usepackage{xcolor}
\definecolor{codegreen}{rgb}{0,0.6,0}
\definecolor{codegray}{rgb}{0.5,0.5,0.5}
\usepackage{float}
\usepackage{listings}
\floatstyle{ruled}
\newfloat{listing}{tb}{lst}{}
\floatname{listing}{Listing}

\usepackage{hyperref}
\usepackage{siunitx}
\usepackage{graphicx}
\newsavebox\CBox
\def\textBF#1{\sbox\CBox{#1}\resizebox{\wd\CBox}{\ht\CBox}{\textbf{#1}}}

\usepackage[accepted]{icml2023}

\usepackage{amsmath}
\usepackage{amssymb}
\usepackage{mathtools}
\usepackage{amsthm}
\usepackage{algorithm}
\usepackage{algorithmic}
\usepackage{float}
\usepackage{caption}
\usepackage{array}
\usepackage{enumitem}
\newcolumntype{?}{!{\vrule width 2pt}}

\DeclareMathOperator*{\argmin}{arg\,min}

\usepackage[capitalize,noabbrev]{cleveref}

\theoremstyle{plain}

\theoremstyle{definition}

\theoremstyle{remark}

\usepackage[textsize=tiny]{todonotes}

\icmltitlerunning{Hierarchical Imitation Learning with Vector Quantized Models}

\begin{document}

\twocolumn[
\icmltitle{Hierarchical Imitation Learning with Vector Quantized Models}




\begin{icmlauthorlist}
\icmlauthor{Kalle Kujanp\"{a}\"{a}}{aaltocs,fcai}
\icmlauthor{Joni Pajarinen}{aaltoelec,fcai}
\icmlauthor{Alexander Ilin}{aaltocs,fcai}
\end{icmlauthorlist}

\icmlaffiliation{aaltocs}{Department of Computer Science, Aalto University, Finland}
\icmlaffiliation{aaltoelec}{Department of Electrical Engineering and Automation, Aalto University, Finland}
\icmlaffiliation{fcai}{Finnish Center for Artificial Intelligence FCAI}

\icmlcorrespondingauthor{Kalle Kujanp\"{a}\"{a}}{kalle.kujanpaa@aalto.fi}

\icmlkeywords{Machine Learning, ICML, Hierarchical RL, Hierarchical Imitation Learning, Reinforcement Learning, Planning, Subgoal Search, Imitation Learning}

\vskip 0.3in
]



\printAffiliationsAndNotice{}  

\begin{abstract}
The ability to plan actions on multiple levels of abstraction enables intelligent agents to solve complex tasks effectively.
However, learning the models for both low and high-level planning from demonstrations has proven challenging, especially with higher-dimensional inputs.
To address this issue, we propose to use reinforcement learning to identify subgoals in expert trajectories by associating the magnitude of the rewards with the predictability of low-level actions given the state and the chosen subgoal.
We build a vector-quantized generative model for the identified subgoals to perform subgoal-level planning.
In experiments, the algorithm excels at solving complex, long-horizon decision-making problems outperforming state-of-the-art.
Because of its ability to plan, our algorithm can find better trajectories than the ones in the training set.
\end{abstract}

\section{Introduction}
\label{introduction}

Learning from expert demonstrations has proven successful in many sequential decision-making settings that can be modeled with Markov decision processes \cite{abbeel2004apprenticeship}. Imitation learning (IL) is a technique for learning to imitate the behavior of an expert by discovering the mapping between states and actions without access to information such as rewards \cite{osa2018algorithmic}.
IL has proven useful in aviation \cite{sammut1992learning}, autonomous driving \cite{chen2022learning}, robotics \cite{kober2008policy}, video games \cite{team2019alphastar} and even healthcare \cite{mayer2008system}.

Recent advances in planning with learned dynamics models have improved our ability to solve complex long-horizon problems when interacting with the environment
is possible \cite{hafner2022deep, ye2021mastering, schrittwieser2020mastering}.
Learning models for planning in the offline setting is possible as well \cite{argenson2020model}, but these model-based reinforcement learning (RL) methods assume access to environment rewards and are not directly applicable to the IL setting.
Hierarchical RL with decision-making at multiple time scales has succeeded in tasks where flat RL struggles \cite{hafner2022deep, nachum2019does}. A hierarchy should also be useful in the IL setting as many real-world problems have a natural hierarchical structure \cite{sharma2019third, jing2021adversarial}. Moreover, planning with a hierarchy may shorten the effective planning horizon and avoid compounding model errors also in the IL setting \cite{nair2019hierarchical}.

We propose a method for hierarchical planning in the IL setting that relies on segmenting expert trajectories into subtasks without any high-level supervision. Unlike prior work that assumes a fixed number of subgoals \cite{pertsch2020keyin}, fixed-length subtasks \cite{czechowski2021subgoal}, or trains multiple models to deal with subtasks of different lengths \cite{zawalski2022fast}, our algorithm segments the trajectories into a variable number of variable-length subtasks. We use the segmentation to learn a generative model over the subgoals and a subgoal-conditioned low-level policy to execute the subtasks. To perform high-level planning, we use standard search methods such as Policy-Guided Heuristic Search \cite{orseau2021policy}, Monte Carlo Tree Search \cite{coulom2006efficient}, or A* \cite{hart1968formal} in which our generative model is used for node expansion.
Our method outperforms strong search, hierarchical IL, and offline RL algorithms at complex long-horizon decision-making problems. Our experiments also show that our algorithm can handle suboptimal expert trajectories and self-improve. 
In summary, the main contributions of this work are:
\begin{enumerate}[noitemsep, topsep=0pt]
    \item A novel yet conceptually simple RL approach for identifying subgoals from trajectories based on the prediction performance of a low-level policy.
    \item A VQVAE generative model for proposing subgoals for planning with temporal abstraction.
    \item In experiments, our generative model combined with the learned low-level policy and a suitable high-level search algorithm solves complex problems with sparse rewards better and with fewer node expansions than state-of-the-art subgoal search and outperforms offline RL algorithms.
\end{enumerate}

\section{Related Work}

Our method combines hierarchical discrete planning with imitation learning to solve complex planning problems. Hierarchical planning allows shortening the effective planning horizon, which is beneficial in long-horizon tasks \cite{pertsch2020long,nair2019hierarchical}. We use offline data to learn a VQVAE \cite{van2017neural} that generates adaptive horizon subgoals and a low-level policy to reach the subgoals for high-level planning.
Previous work proposes a wide variety of different approaches for
learning hierarchical behavior from data, also in combination with
planning. However, according to our knowledge, our method is the first method that segments a trajectory into a varying number of adaptive-length subtrajectories and uses the subgoals to learn a generative model that can solve difficult long-term decision-making problems in a discrete setting.

\textbf{Hierarchical RL.} \citet{sutton1999between} proposed using options, a set of low-level policies with termination, for decision-making at a higher level of temporal abstraction than standard RL. Options can be learned end-to-end without supervision \cite{bacon2017option, bagaria2020option, sharma2019dynamics} and applied to planning \cite{silver2012compositional}. Representing the hierarchy as subgoals is an alternative to options \cite{dayan1992feudal}, which is the approach we adopt in our work. Our work differs from these methods in that we assume that we cannot interact with the environment and must learn the representation from a given dataset.

\textbf{Hierarchical Continuous Planning.}
In hierarchical planning with continuous control, the Cross-Entropy Method (CEM) is typically used.
With continuous controls, search methods with convergence guarantees such as MCTS \cite{coulom2006efficient}, PHS$^*$ \cite{orseau2021policy}, or A$^*$ \cite{hart1968formal}, utilized with our method in the experiments, cannot be directly applied. HiGoC \cite{li2022hierarchical} uses CEM for hierarchical planning in offline RL assuming access to rewards. HVF \cite{nair2019hierarchical} adds a hierarchical structure with predicted subgoal images to Visual Model Predictive Control \cite{ebert2018visual} that plans in image space using CEM. Visual hierarchical methods that rely on identifying keyframes from trajectories using variational inference and planning with CEM include TAP \cite{jayaraman2018time} and KeyIn \cite{pertsch2020keyin}.
However, these methods assume a fixed number of keyframes in the trajectories and have not been successfully applied to highly complex reasoning tasks with sparse rewards. KeyIn also uses an environment simulator for planning.

Goal-Conditioned Hierarchical Planning \cite{pertsch2020long} 
produces plans, executed by a learned inverse dynamics model, in a high-dimensional state space in an offline setting.
SGT-PT does goal-based RL in a low-dimensional setting by planning with subgoal trees \cite{jurgenson2020sub}. 
However, these models require an explicit goal state and need to generate subgoals between the initial state and the goal state, which can be a difficult learning problem in complex long-horizon environments.

\textbf{Hierarchical IL without Planning.}
Unlike our method, many model-free hierarchical imitation learning methods assume some degree of high-level supervision \cite{le2018hierarchical,fox2019multi,zhang2021explainable}.
As an alternative \citet{daniel2016probabilistic} infer compositional structure in data discovering options 
with expectation-maximization. CompILE \cite{kipf2019compile} uses VAEs to segment trajectories into subtasks and the subtask encodings as subpolicies in hierarchical RL. The model-free method Option-GAIL \cite{jing2021adversarial} infers expert options from trajectories with an EM-like approach. \citet{zhang2021provable} 
directly optimize a hierarchical policy with options by maximizing the probability of expert trajectories with a hidden Markov model.
Directed-Info GAIL \cite{sharma2018directed} is a variant of hierarchical inverse RL that learns latent policies by modeling problems as directed graphs. The method segments expert trajectories into sub-tasks and learns structural policies to solve different sub-tasks. OptionGAN \cite{henderson2018optiongan} learns to recover reward and policy options simultaneously. Learning from Guided Play (LfGP) \cite{ablett2021learning} uses scheduled auxiliary tasks to address lacking exploration in adversarial online IL. \citet{paul2019learning} learn a generative model over subgoals from demonstrations and use it to augment the reward function for RL fine-tuning. However, these methods do not incorporate high-level planning mechanisms, which may make them unsuitable for solving complex reasoning problems.

\begin{figure*}[t]
\centering
\includegraphics[width=\textwidth]{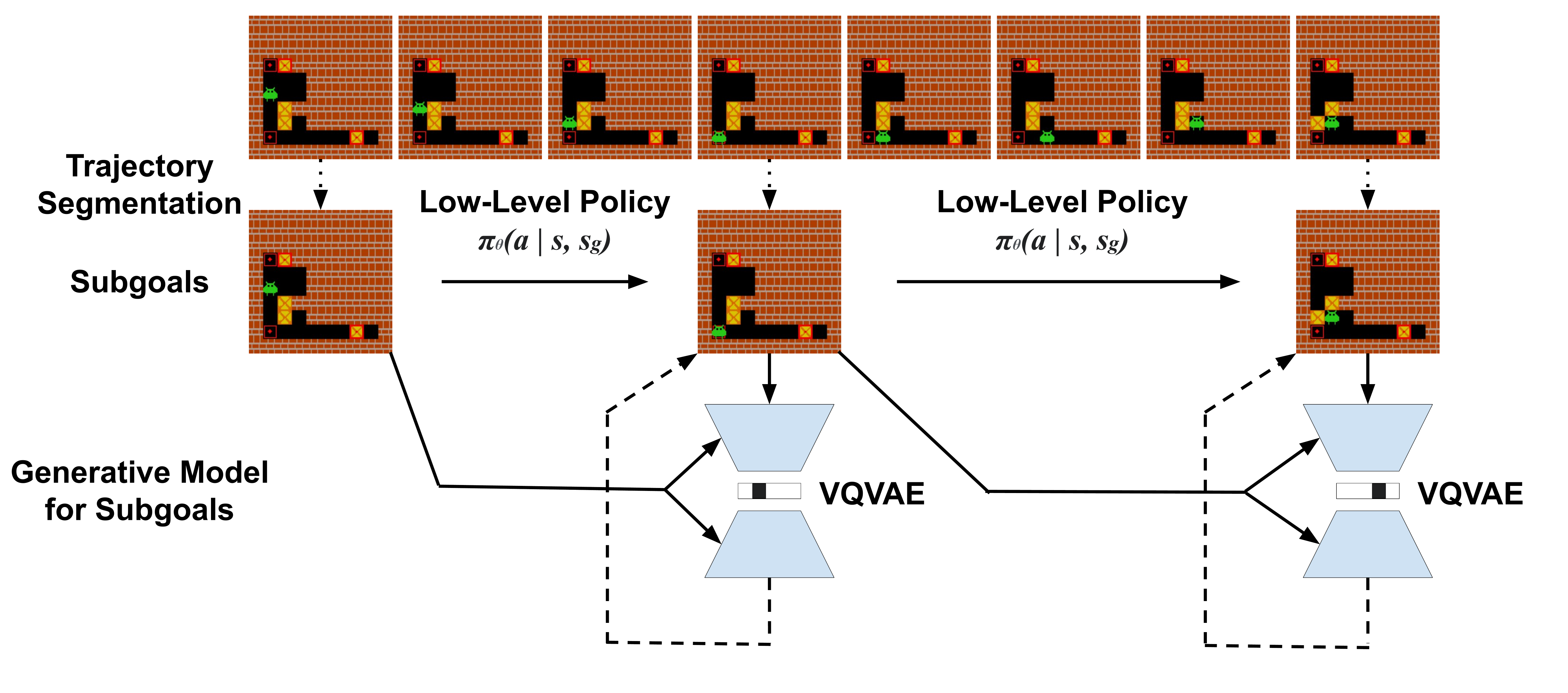}
\caption{A visualization of our \textit{Hierarchical Imitation Planning with Search} (HIPS) when learning to solve Sokoban. The main components of our method are a detector $d_{\xi}(s_{g_k} | s_i)$ that segments the trajectory into subgoals, a subgoal-conditioned low-level policy $\pi_\theta(a_i | s_i, s_{g_k})$, and the VQVAE, a generative model over the subgoals. The low-level policy and VQVAE are used during evaluation for planning, whereas the detector is training-only.}
\label{fig:hips}
\end{figure*}

\textbf{Offline RL.}
In offline RL, the agent's objective is to learn an optimal policy without interacting with the environment. Instead, the agent has access to a dataset of transitions that have been collected by a behavior policy $\pi_\beta$ that can be suboptimal. A significant benefit that offline RL has over imitation learning is the ability to extract strong policies even when the expert trajectories are suboptimal \cite{kumar2022should}. Conservative Q-learning \cite{kumar2020conservative} is a model-free offline RL method that learns a lower bound on the policy value, which helps avoid overestimating state values. Decision Transformer \cite{chen2021decision} treats the offline reinforcement learning task as a sequence modeling problem, where the goal is to predict the action conditioned on a desired reward. We use offline RL methods as baselines in our experiments.

\textbf{Hierarchical IL with Discrete Search.}
The closest works to ours in search and planning are kSubS \cite{czechowski2021subgoal} and AdaSubS 
\cite{zawalski2022fast}. kSubS learns an autoregressive model for generating subgoals given a set of trajectories and plans with subgoal search.
Unlike our method, kSubS relies
on the true environment dynamics for low-level search to find fixed-length subtrajectories 
between subgoals in some environments when combined with an autoregressive CNN. AdaSubS replaces the low-level search of kSubS with a learned policy and supports subtrajectories of multiple hard-coded lengths by training an autoregressive generative model for each length. Our approach relies on a learned low-level policy and dynamics model and supports varying-length segments. We show that it is possible to train a single non-autoregressive adaptive generative model, which makes generating subgoals more efficient.

\section{Method}

We consider goal-oriented, complex reasoning problems, in which the agent's objective is to act in the environment to reach a terminal state. This corresponds to a Markov decision process with a reward of one upon solving the task and zero otherwise. We consider environments with fully Markovian, discrete-valued states with full observability.
We work in the imitation learning (offline) setting: the agent needs to learn to solve tasks only from available demonstrations without the possibility to interact with the environment before evaluation. We assume that there is a dataset $\mathcal{D}$ of trajectories $\tau = \{s_0, a_0, s_1, \dots, a_{T-1}, s_T\}$ collected by experts who know how to solve tasks, with only a reward of one at the terminal state $s_T$. The experts may not reach the terminal states in the fastest possible way. Our method should be able to combine parts of the dataset trajectories, that is, perform stitching \citep{singh2020cog}, to discover efficient solutions.

We propose to solve the imitation learning task using an agent which has a hierarchical structure with two levels. 
Our agent learns a hierarchical representation of the available trajectories by identifying likely experts' subgoals in the existing trajectories in an unsupervised manner. A low-level policy for reaching these subgoals is learned simultaneously. The identified subgoals are then used to train a discrete-code generative model which can generate reasonable subgoals to perform subgoal-level planning with standard search algorithms such as PHS, MCTS, or A*. The low-level policy executes the plan generated by the planner by sequentially reaching the determined subgoals. Planning with search allows our method to improve on suboptimal demonstrations. A graphical representation of our method is shown in Fig.~\ref{fig:hips}. We call our method \textit{Hierarchical Imitation Planning with Search} (HIPS). Below we describe the main components of the approach: the subgoal detector $d_\xi(s_{g_{k+1}} | s_{g_k})$, the subgoal conditioned low-level policy $\pi_\theta(a_i | s_i, s_{g_k})$, and the subgoal generative model $p(s_g | s)$.

\subsection{Subgoal Identification}

The goal of the subgoal identification phase is to learn a high-level representation $\tau_* = (s_{g_1}, s_{g_2}, \dots, s_{g_M})$ of each trajectory such that the trajectory is represented as a sequence of subgoals $s_{g_k}$. Each subgoal is a state from the trajectory, that is $\forall_k s_{g_k} \in \tau$, and in particular, $s_{g_M} = s_T$. This is a time series segmentation problem where the solution has three desirable qualities: 1)~we want the identified subgoals $s_{g_k}$ to be easy to reach by a trainable low-level policy $\pi_{\theta}(a_i | s_i, s_{g_k})$ which takes the subgoal $s_{g_k}$ as input, 2)~we want the subgoals to be easy to sample from a generative model, and 3) we want to segment the trajectories into as few subgoals as possible to make planning over them more efficient.
Successfully segmenting the trajectories into a variable number of variable-length segments turns out to be a highly non-trivial task, as most prior work has focused on fixed-length segments or a fixed number of segments.
Our solution is to formulate this segmentation task as an RL problem in which we treat each trajectory from $\mathcal{D}$ as one episode for training, and try to maximize the subgoal-conditioned log-probability of the actions in the trajectory, $\sum_{i=0}^{T-1} \log \pi_\theta(a_i | s_i, s_{g_k})$, and minimize the number of segments.

In each episode, the segmentation agent starts at the first state $s_0$ of the trajectory and selects the next subgoal state $s_{g_1}$ according to the probabilities produced by its policy $d_{\xi}(s_{g_1} | s_0)$, which we call \emph{the detector}. We limit the detector to consider only the following $H$ states as candidate subgoals, that is $s_{g_1}$ is selected from $s_1, ..., s_H$. The agent samples the next subgoal according to the computed probabilities and gets the reward
\begin{align}
    R_1 = r_{1} -\alpha,
\label{eq:reward}
\end{align}
where $r_1$ is the log-probability that a low-level policy $\pi_\theta(a_i | s_i, s_{g_1})$ selects the sequence of actions $a_0, ..., a_{g_1-1}$ in the first segment:
\begin{align*}
    r_1 =  \sum_{i=0}^{g_1-1}\log \pi_\theta(a_i | s_i, s_{g_1})
\end{align*}
and $\alpha$ is a penalty to prevent segmentation into too many subtrajectories.
After that, the segmentation agent changes its state to $s_{g_1}$, selects the next subgoal according to $d_\xi(s_{g_2} | s_{g_1})$ and gets reward $R_2$ computed using action log-probabilities from the second segment, similarly to \eqref{eq:reward}. The episode continues like this until the end of the trajectory is reached.

The low-level policy $\pi_\theta(a_i | s_i, s_{g_k})$ is considered as part of the environment of the segmentation agent. It is updated during training using goal-conditioned behavioral cloning to minimize
\begin{align}
    \mathcal{L_\theta} = - \mathbb{E}_{\tau \sim \mathcal{D}} \sum_{k=1}^{M(\tau)} \sum_{i=g_{k-1}}^{-1 + g_k} \log \pi_\theta(a_i | s_i, s_{g_k}),
\label{eq:gcbc}
\end{align}
with subgoals $s_{g_k}$ produced by the segmentation agent. Note that the number of identified subgoals $M(\tau)$ may vary across trajectories.

Thus, the segmentation agent is trained by giving it a higher reward when selected subgoals lead to more accurate action predictions by the low-level policy. The low-level policy is trained concurrently with the subgoals (high-level commands) produced by the segmentation agent as input. We train the segmentation agent using the policy gradient algorithm REINFORCE with a learnable value function as baseline \cite{williams1992simple}.

We train the low-level policy and detector simultaneously, and the associated non-stationarity might cause issues. In practice, we observe that $\pi_\theta(a_i | s_i, s_{g_k})$ converges rapidly during the simultaneous training and its effect on the non-stationarity is limited. Note that using a trainable detector $d_\xi(s_{g_{k+1}} | s_{g_k})$ may encourage subgoals that are easy to recognize among the states in the training trajectories. If that is true, we hypothesize that such subgoals may be easy to produce by a learned generative model. Our approach for segmenting trajectories is summarized in Algorithm~\ref{alg:segment}.

\begin{algorithm}[tb]
\caption{Segmenting Trajectories for Hierarchical IL}
\label{alg:segment}
\textbf{Input}: A dataset of trajectories $\mathcal{D}$, untrained low-level policy network $\pi_\theta$, detector $d_\xi$ \\ 
\textbf{Parameters}: The parameters of the low-level policy network, $\theta$, detector network, $\xi$ \\ 
\textbf{Output}: Trained low-level policy $\pi$, dataset $\mathcal{D}'$ of subgoal pairs $\{(s_{g_{k+1}}, s_{g_{k}})\}$ 
\begin{algorithmic}[1] 
\WHILE{$\pi_\theta$, $d_\xi$ not converged} 
\STATE Sample a trajectory $\tau$.
\STATE Segment $\tau$ with $d_\xi$
\STATE Predict low-level actions with $\pi_\theta$ conditioned on produced subgoals
\STATE Compute returns (Equation~\ref{eq:reward}), update $\xi$ with REINFORCE 
\STATE Compute the losses for $\pi_\theta$ (Equation~\ref{eq:gcbc}), update $\theta$  
\ENDWHILE
\STATE Create a dataset $\mathcal{D}'$ of subgoal pairs $\{(s_{g_{k+1}}, s_{g_{k}})\}$ by sampling trajectories $\tau$ from $\mathcal{D}$ and segmenting them with $d_\xi$.
\STATE \textbf{return} $\pi_\theta, \mathcal{D}'$ 
\end{algorithmic}
\end{algorithm}

\subsection{Generative Model for Subgoals}

To plan in terms of the high-level subgoals, the agent needs the ability to generate reasonable subgoals $s_g$ for each environment state $s$. We do this by learning a generative model $p(s_g | s)$ over the subgoals using the ones identified in the trajectory segmentation step as training data. We implement the model as a VQVAE with discrete latent codes, which is inspired by the vector quantized models proposed by \citet{ozair2021vector}. 
The VQVAE encoder 
takes a pair of states $(s_g, s)$ as input and outputs a continuous latent code $z_e$. Then, the code is quantized by finding the nearest code $e_k$ from the codebook such that $k~=~\argmin_m \| z_e - e_m\|_2$. The decoder 
uses the code $e_k$ to reconstruct the subgoal state $\hat{s}_{g} = g_{\psi}(e_k, s)$. The loss minimized during training is
 \begin{multline}
     \mathcal{L} = 
     \mathcal{L}_\text{rec}(\hat{s}_g, s_g)
     + \| [z_e] - e_k \|_2^2 + \beta \| z_e - [e_k]\|_2^2
\,,
\label{eq:vqvae_loss}
\end{multline}
where $\mathcal{L}_\text{rec}$ is the reconstruction loss and $[\cdot]$ denotes the stop gradient operation.

We train the VQVAE in two stages. In the pre-training stage, we use random pairs of states $(s_j, s_i), i < j \leq i+H$ from the trajectories as inputs $(s_g, s)$. 
We skip the discretization layer
and only use the reconstruction loss $\mathcal{L}_\text{rec}(\hat{s}_g, s_g)$ 
for training the encoder and the decoder. After the pre-training has converged, the complete VQVAE is trained by using pairs of consecutive subgoals $(s_{g_{k+1}}, s_{g_{k}})$ as input. We initialize the codebook by running KMeans++ clustering \cite{arthur2006k} on the first batches of encoder outputs and using the cluster centers as the initial codes. This training strategy was inspired by the strategy proposed by \citet{lancucki2020robust}.
Finally, we learn a subgoal-conditioned prior $p(e_{k+1} | s_{g_{k}})$ over the latent codes.

Once the model has been trained, one can generate subgoals conditioned on the current state $s$ by sampling a code $e$ from the learned codebook and running it through the decoder $g_\psi(e, s)$. Note that the number of possible codes $e$ is finite, which means that the number of generated subgoals $s_g$ is finite as well.
Note also that distinct codes $e$ may result in the same generated subgoal $s_g$, which is a desired behavior when the size of the codebook is larger than the number of reasonable subgoals for the considered state. The pseudo-code for our VQVAE training is given in Algorithm~\ref{alg:vqvae} in the Appendix~\ref{app:vqvae}. 

\begin{figure*}[t]
\centering
\begin{minipage}{.24\linewidth}
\centering
\includegraphics[height=40mm,trim={0mm 2mm 210mm 3mm},clip]{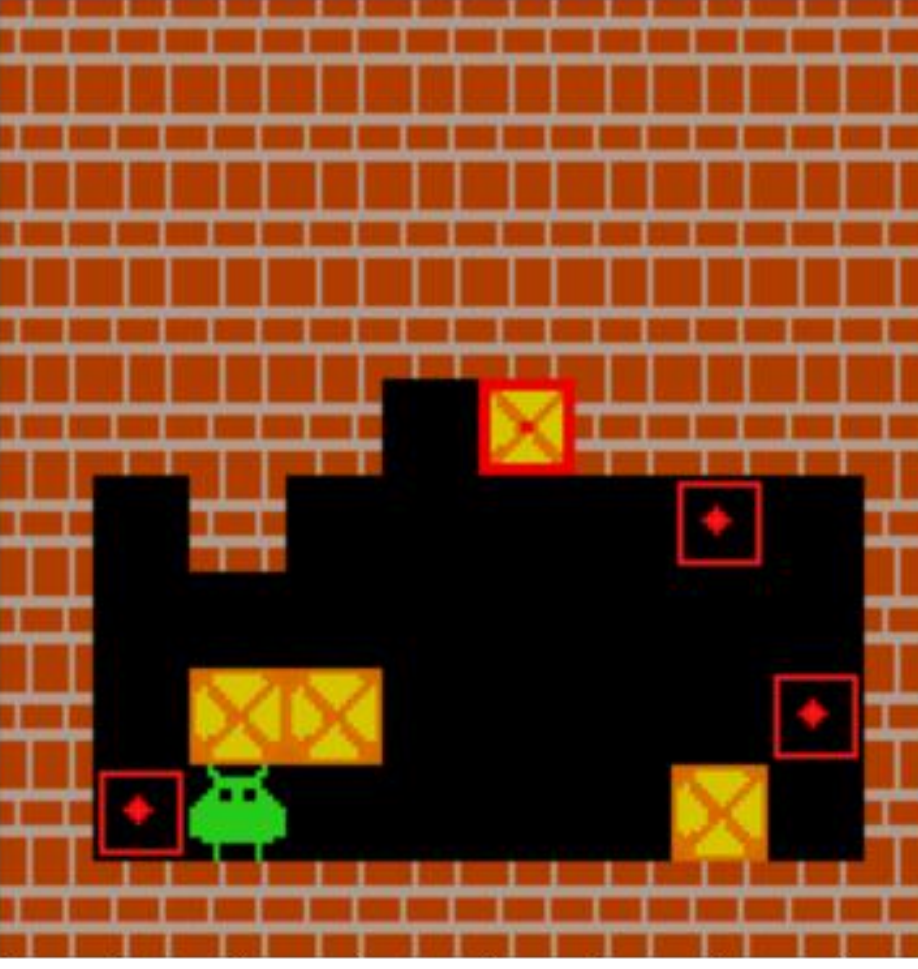}
\\
Sokoban
\end{minipage}
\hfil
\begin{minipage}{.24\linewidth}
\centering
\includegraphics[height=40mm,trim={100mm 0 100mm 0},clip]{Figures/AllEnvs.pdf}
\\
Sliding Tile Puzzle
\end{minipage}
\hfil
\begin{minipage}{.24\linewidth}
\centering
\includegraphics[height=40mm,trim={175mm 21mm 175mm 21mm},clip]{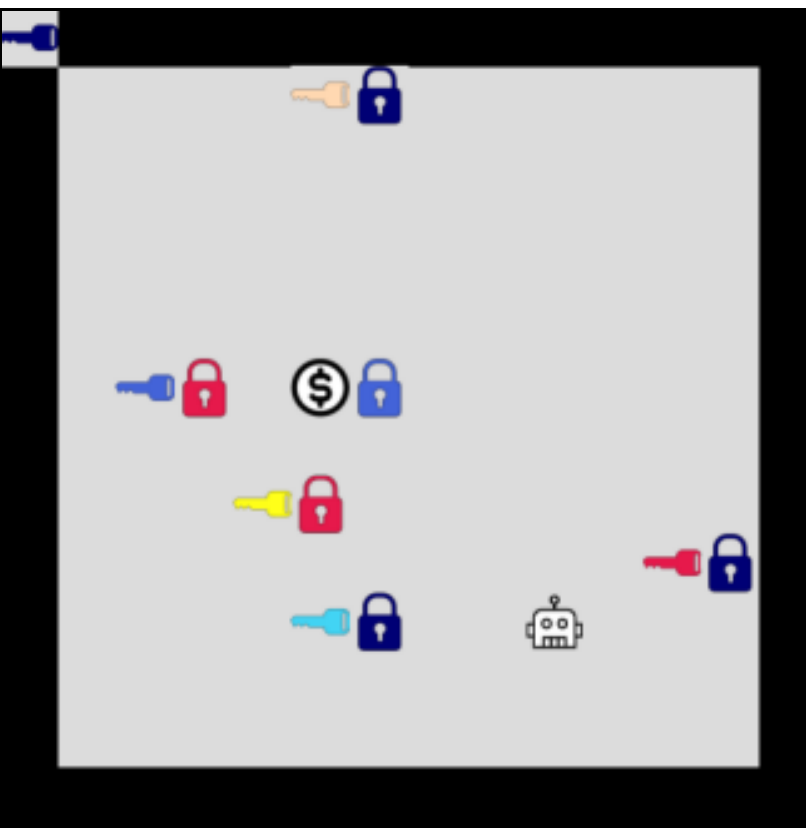}
\\
Box-World
\end{minipage}
\hfil
\begin{minipage}{.24\linewidth}
\centering
\includegraphics[height=40mm,trim={85mm 22.5mm 495mm 22.5mm},clip]{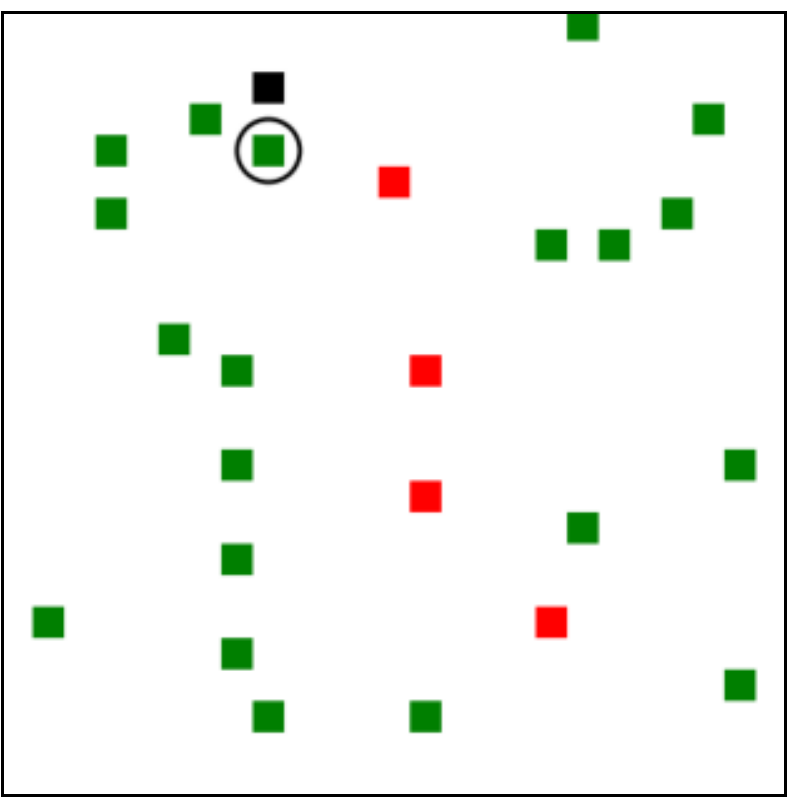}
\\
Traveling Salesman Problem
\end{minipage}

\caption{The environments we consider in the experiments.
Sokoban: the task is to push yellow boxes onto the target locations (marked with red squares).
Sliding Tile Puzzle (STP): The task is to order the tiles from 1 to 24 by moving them.
Box-World:
The agent must collect colored keys and open color-matching locks to recover more keys until it finally reaches a goal target (marked with the \$ sign).
Traveling salesman problem (TSP): The agent (marked with the circle) has to visit all cities (marked with squares) before returning to the start (marked with the black square). Visited cities are marked with green squares and unvisited ones with red squares.}
\label{fig:env_illustration}
\end{figure*}

\subsection{High-Level Planning with Search}

We perform planning in the subgoal space, as it can be more efficient and suitable for long-horizon planning than planning in the state space \cite{nair2019hierarchical}. We demonstrate that our method is compatible with many different search algorithms. 
Each search node represents a subgoal. When a search node is expanded, possible next subgoals (child nodes) are generated with the VQVAE. Given a codebook of size $K$, there are $K$ possible child nodes for each subgoal. This can be interpreted as the latent codes encoding the different branches of the search tree rather than the actual subgoals, which helps us significantly constrain the codebook sizes. To limit the size of the search tree, we remove duplicates and unreachable subgoals.
The reachability of a proposed subgoal $s_{g}$ from state $s_i$ is evaluated by using the low-level policy $\pi_\theta(a_i | s_i, s_{g})$ trained in the segmentation phase. We run the policy iteratively from state $s_i$ and simulate state transitions by using the true dynamics or a learned model $f_\text{dyn}({s_{i+1} | a_i, s_i})$. If the subgoal $s_{g}$ is reached within a specific number of steps, the subgoal is considered reachable, and a search node is created. In this work, we work with discrete states and require an exact match between the reached state and the subgoal. Using a suitable threshold to evaluate the match is an alternative in the continuous setting. We also learn a value function $V(s)$ that predicts the number of low-level steps necessary to reach the goal (terminal state) and use it as a heuristic in planning.

The search methods we use are Greedy Best-First Search (GBFS), Policy-Guided Heuristic Search \citep[PHS*,][]{orseau2021policy}, A* \cite{hart1968formal} and Monte-Carlo Tree Search \citep[MCTS,][]{coulom2006efficient, kocsis2006bandit}. PHS* is dependent on a good policy, but when the dataset contains suboptimal trajectories, learning a good VQVAE prior to act as the policy might be impossible. Then, policy-independent algorithms like A* or GBFS can be superior. PHS* is also aimed at minimizing the search loss, not finding a particularly high-quality solution. When that is important, A* or MCTS can be superior to PHS*.

\section{Experiments}

In our experimental phase, we evaluate our method on complex, sparse reward problems that require reasoning. We compare our method to existing search, hierarchical imitation learning, and offline RL methods. We also analyze whether our RL-based approach for identifying subgoals is superior to subgoals sampled at fixed intervals.

\subsection{Environments}

We evaluate our method in four environments that are all complex reasoning domains (see Fig.~\ref{fig:env_illustration}). The first environment is Sokoban, which is a PSPACE-complete puzzle where the agent must push boxes onto goal locations \cite{culberson1997sokoban}. The moves are irreversible and one wrong push can make the puzzle unsolvable. We use a $10 \times 10$ problem size with four boxes, the default configuration in the earlier literature \cite{orseau2021policy, guez2019investigation, racaniere2017imagination}. We use a one-hot encoded tensor with shape $10 \times 10 \times 4$ as the observation space \cite{orseau2021policy}.

The second environment is the sliding tile puzzle (STP) which is a classic benchmark in the search literature \cite{korf1985depth}. We use a puzzle size of $5 \times 5$, and the objective is to sort the number tiles in a specific descending order.

The third environment is Box-World (BW), where the agent must collect colored keys and open color-matching locks to recover more keys until it finally reaches a goal target \cite{zambaldi2018relational}. Keys can only be used once. If the agent uses its key to open the wrong box, the game will become unsolvable. Hence, careful planning and reasoning about entities and their relations are required.

The fourth environment is a grid-based Traveling Salesman Problem (TSP). The agent moves through the 2D grid to visit all the cities before returning to the starting point (see Fig.~\ref{fig:env_illustration}). The TSP is an NP-hard combinatorial optimization problem. However, finding any solution to TSP is relatively easy. Hence, grid-based TSP serves as an environment for evaluating the solution quality.

In Sokoban, we collect a training set of 10340 trajectories using gym-sokoban \cite{SchraderSokoban2018}. 10240 of the problem instances have been solved using Curry \cite{curry} and 100 trajectories were collected by performing random actions. The training set consists of 5100 trajectories in STP and 22100 in Box-World, of which 5000 and 22000 were collected by solving the problem instances with a subgoal-based A* algorithm \cite{hart1968formal} and 100 by performing random actions. In subgoal-based A*, we generated subgoals progressively closer to the terminal state procedurally and executed A* to reach these subgoals, as solving complete problem instances with A* would have been computationally very expensive due to the complexity of the environments. In TSP, we do not limit the number of demonstrations available to the agent but the demonstration trajectories are highly suboptimal and generated by running an agent that visits 25 cities in a $25 \times 25$ grid in random order.

\subsection{Agents}

Our HIPS agent consists of the following neural networks:
the detector $d_\xi(s_{g_{k+1}} | s_{g_{k}})$,
the low-level policy $\pi_\theta(a_i | s_i, s_{g_k})$,
the VQVAE encoder $f_\phi(z_{e_{k+1}} | s_{g_{k+1}}, s_{g_{k}})$,
the VQVAE decoder $g_\psi(s_{g_{k+1}} | e_{k+1}, s_{g_k})$,
the VQVAE prior $p(e_{k+1} | s_{g_{k}})$,
the low-level dynamics model $f_\text{dyn}({s_{i+1} | a_i, s_i})$, and
the distance function $V(s_i)$.
The encoder $f_\phi$ and detector $d_\xi$ are not used during evaluation. All networks are ResNet-based CNNs \cite{he2016deep}. The decoder and low-level dynamics CNNs also contain FiLM layers \cite{perez2018film}. We only use the 100 random trajectories for training the low-level dynamics model. In Box-World, the neural networks additionally use Deep Recurrent Convolutions \cite{guez2019investigation}. All neural networks are implemented with PyTorch \cite{paszke2019pytorch} and trained with an Adam optimizer \cite{kingma2014adam}. We evaluate Sokoban and Box-World with PHS* as the high-level search algorithm. Because of the suboptimality of the TSP trajectories, learning a good VQVAE prior is difficult, and we use A* in TSP. We also observed that GBFS is superior to PHS* with very small search budgets in STP (see Appendix~\ref{app:k}).

\begin{table}[t]
\caption{The overall success rates (\%) of different algorithms. The algorithms in the bottom part have access to the true environment dynamics, and those in the upper part do not.}
\centering
\begin{tabular}{lrrrr}
Method & Sokoban & STP & BW & TSP\\
\midrule
HIPS (ours) & 97.5 & \textBF{94.7} & \textBF{55.7} & \textBF{100.0} \\
HIPS-$k$ (ours) & \textBF{99.0} & \textbf{94.7} & 20.1 & \textBF{100.0} \\
BC & 18.7 & 82.5 & 41.1 & 28.8 \\
CQL & 3.3 & 11.7 & 6.0 & 33.6\\
DT & 36.7 & 0.0 & 10.0 & 0.0\\
RIS & 0.0 & 0.0 & 7.3 & 0.0 \\
Option-GAIL & 0.3 & 0.0 & 0.0 & 0.0\\
IQ-Learn & 0.0 & 0.0 & 0.0 & 0.0\\
\midrule
HIPS-env (ours) & 98.1 & 94.6 & 99.6 & 100.0 \\ %
AdaSubS & 91.4 & 0.0 & & 22.4 \\
kSubS & 90.5 & 93.3 & & 87.9\\
\end{tabular}
\label{tab:solutionRate}
\end{table}

\begin{table*}[t]
\caption{The success rates (\%, higher is better) of different search algorithms after performing $N$ node expansions. In Sokoban and Box-World, HIPS was evaluated with PHS*, with GBFS in STP, and with A* in TSP.}
\centering
\begin{tabular}{l|rrr|rrr|rrr|rrr}
\multicolumn{1}{c|}{} & \multicolumn{3}{c|}{\textbf{Sokoban}} & \multicolumn{3}{c|}{\textbf{Sliding Tile Puzzle}}  & \multicolumn{3}{c|}{\textbf{Box-World}} &  \multicolumn{3}{c}{\textbf{Travelling Salesman}}\\
\midrule
$N$ & 100 & 500 & 1000 & 100 & 500 & 1000 & 100 & 500 & 1000 & 100 & 500 & 1000 \\
\midrule
HIPS (ours) & \textbf{88.5} & 94.7 & 95.9 & 89.8 & 91.8 & 92.2 & 55.7 & 55.7 & 55.7 & \textbf{100.0} & \textbf{100.0} & \textbf{100.0} \\
HIPS-$k$ (ours) & 77.7 & 90.9 & 94.3 & 68.0 & 79.1 & 84.4 & 20.1 & 20.1 & 20.1 & 77.3 & \textbf{100.0} & \textbf{100.0}\\
\midrule
HIPS-env (ours) & \textBF{88.5} & \textBF{94.9} & \textBF{96.4} & \textBF{90.2} & \textBF{92.1} & 92.6 & \textBF{99.6} & \textBF{99.6} & \textBF{99.6} & \textBF{100.0} & \textBF{100.0} & \textBF{100.0} \\
HIPS-env-$k$ (ours) & 76.9 & 91.3 & 94.0 & 69.7 & 80.9 & 87.2 & 99.1 & 99.1 & 99.1 & 77.2 & \textbf{100.0} & \textbf{100.0} \\
AdaSubS & 82.2 & 88.8 & 90.2 & 0.0 & 0.0 & 0.0 & & & & 1.2 & 12.2 & 20.8 \\ 
kSubS & 73.1 & 79.9 & 82.8 & 79.9 & 91.7 & \textbf{92.7} & & & & 40.4 & 80.9 & 85.3 \\
PHS* & 1.8 & 76.1 & 91.1 & 0.0 & 0.0 & 0.0 & 25.8 & 93.8 & 94.0 & 0.0 & 0.0 & 0.0 \\
\end{tabular}
\label{tab:searchExpansions}
\end{table*}

We compare our agents to two main classes of baselines.
The first class of baselines is strong IL and offline RL algorithms. We compare our agent to standard flat behavioral cloning (BC), a powerful offline RL algorithm, Conservative Q-Learning \citep[CQL,][]{kumar2020conservative}, and a strong IL algorithm, Inverse Soft-Q Learning \citep[IQ-Learn,][]{garg2021iq}. We ran IQ-Learn in the online mode, where it could collect additional data during training. Furthermore, we include the Decision Transformer \citep[DT,][]{chen2021decision}, the hierarchical IL algorithm Option-GAIL \cite{jing2021adversarial}, and the online goal-conditioned RL algorithm RIS \cite{chane2021goal} in our experiments. For CQL, we use the implementation of d3rlpy \cite{seno2021d3rlpy}, and for other algorithms, we use the open-sourced implementations of the authors. For all methods that need rewards, we give the agent a reward of one upon completing the task and 0 otherwise. We do not include the 100 random trajectories in the dataset when evaluating imitation learning algorithms. 

The second class of baselines is search methods that use the true environment dynamics, a state-of-the-art low-level search, Policy Guided Heuristic Search \citep[PHS*,][]{orseau2021policy}, and two subgoal-level search methods: kSubS \cite{czechowski2021subgoal} and AdaSubS \cite{zawalski2022fast}. In PHS*, we observed that using a policy trained with behavioral cloning works better than a policy trained using the loss function proposed by \citet{orseau2021policy} and therefore, we use the BC policy in the experiments. We evaluate kSubS and AdaSubS with the autoregressive CNNs on all environments except Box-World because it would have required significant changes to the existing implementation.
Our method, HIPS, relies on a learned dynamics model instead of the true dynamics. Hence, it solves a more complex problem. We also train a more comparable variant of our method, HIPS-env, that uses an environment simulator for planning.

\subsection{Results}

We use the overall success rates reported in Table~\ref{tab:solutionRate} as the main evaluation metric in all environments. We evaluate the performance 
of each seed and take the mean over the random seeds. We use 10 random seeds to evaluate our method, HIPS, at least five seeds per ablation, and at least three seeds per baseline method. When evaluating the overall success rate, the search algorithms may perform as many expansions as necessary to find a solution to the problems. The critical success factor for our model is the capacity of the generative model to cover the complete search space with the proposed subgoals.

Our method, HIPS, outperforms the baseline methods in all four environments (see Table~\ref{tab:solutionRate}). 
The table also contains an ablation of our method, HIPS-$k$, where we train the VQVAE with subgoals sampled at fixed intervals as done in AdaSubS and kSubS instead of using the detector network (see Appendices~\ref{app:k}~and~\ref{app:hyperparameters} for more details). Eliminating the detector leads to a clear drop in performance in one of the four environments. HIPS-$k$ is superior to kSubS in all environments despite solving a more difficult problem than kSubS. HIPS-$k$ must learn the environment dynamics and a low-level policy, whereas kSubS uses a low-level search with the environment dynamics. HIPS-$k$ also is superior to AdaSubS which also uses a low-level policy instead of search. The sparse reward structure and the required reasoning capabilities prove to be very difficult for the model-free baseline IL and offline RL methods that do not rely on planning, which is why they struggle with all four tasks.

\begin{table}[t]
\caption{The success rates of different algorithms (higher is better) and the average number of steps (lower is better) needed to solve TSP.}
\centering
\begin{tabular}{lrr}
 Method & Success rate (\%) & Avg. steps \\ 
 \midrule
HIPS-PHS* (ours) & \textBF{100.0} & 305.9 \\ 
 HIPS-MCTS (ours) & \textBF{100.0} & 213.0  \\ 
 HIPS-A* (ours) & \textBF{100.0} & \textBF{168.2} \\
 kSubS & 87.9 & 268.2 \\
 CQL & 33.6 & 336.9 \\ 
 BC & 28.8 & 339.3 \\ 
 AdaSubS & 22.4 & 338.9 \\
 \midrule
 Teacher & 100.0 & 336.5 \\
 Oracle MCTS & 100.0 & 199.9 \\
 Christofides & 100.0 & 139.0 \\
\end{tabular}
\label{tab:solutionRateTSP}
\end{table}

HIPS-env performs equally to HIPS, except in Box-World, where the search exploits the inaccuracies of the dynamics model. However, HIPS was evaluated using open-loop planning, where one plan was generated, executed, and evaluated. If the agent is allowed to re-plan when the dynamics model deviates from the environment and fine-tune the model with the new transition, the performance would most likely increase. The issues with the dynamics model do not prevent HIPS from outperforming the baselines.

\begin{figure*}[t]
\centering
\includegraphics[width=\textwidth,trim={0mm 21mm 0mm 21mm},clip]{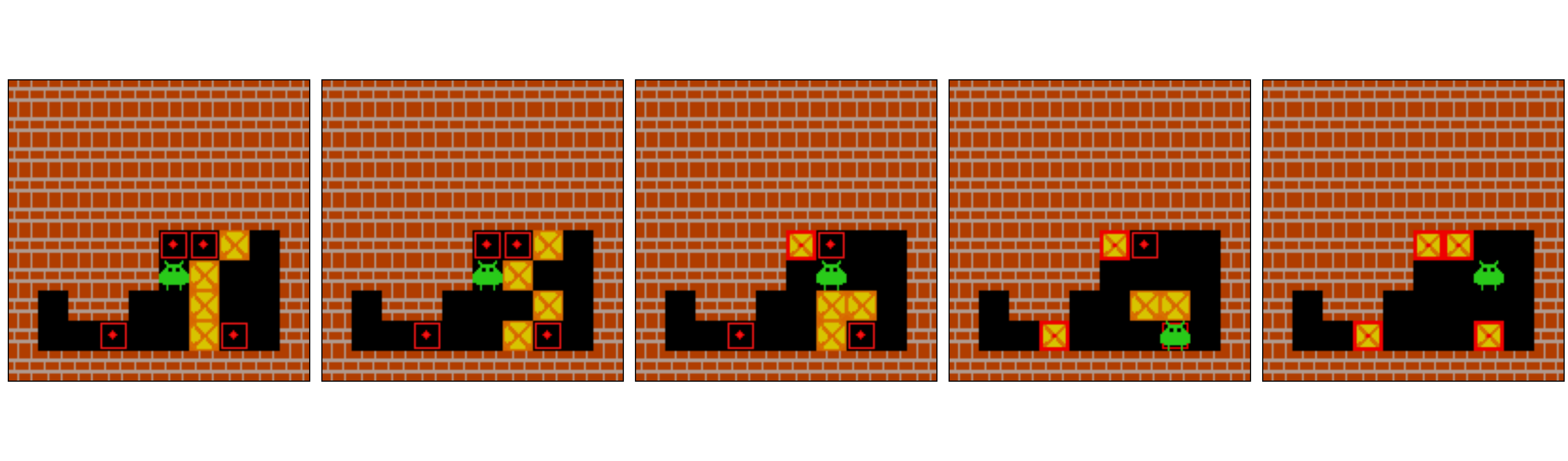}

\caption{An example of a high-level plan (a sequence of subgoals) found by HIPS in Sokoban.}
\label{fig:sokoban_subgoals_text}
\end{figure*}

\begin{figure*}[t]
\centering

\includegraphics[width=\textwidth,trim={0mm 192mm 0mm 0mm},clip]{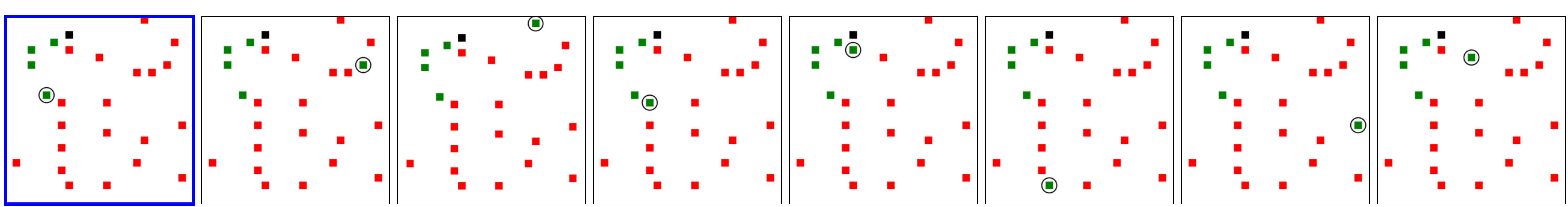}

\caption{
Examples of subgoals proposed for the current state (marked with blue boundaries) in TSP.
}
\label{fig:tsp_subgoals_short}
\end{figure*}

Letting a search algorithm perform unlimited expansions is unrealistic in most real-world applications. Following \citet{czechowski2021subgoal}, we evaluate the percentage of test problems solved after $N$ node expansions. The results are given in Table~\ref{tab:searchExpansions}. The benefits of using RL to detect subframes become clear, as HIPS outperforms the fixed-length ablation HIPS-$k$ in Sokoban, Sliding Tile Puzzle (STP), and Travelling Salesman Problem (TSP). HIPS-env outperforms the baseline methods in Sokoban and TSP and is superior to kSubS on STP when the search budget is small. PHS* can solve STP and TSP, but cannot make enough progress in 1000 node expansions. AdaSubS cannot solve STP because it struggles to reliably reach the generated subgoals with its low-level policy. Therefore, kSubS outperforms AdaSubS. 

TSP is a problem where generating a successful trajectory is easy, but finding an efficient solution is much more difficult. Hence, we evaluate the solution lengths of the methods. We also compare the search algorithms PHS*, MCTS, and A* when used with HIPS. Given the amount of data, we perform VQVAE pre-training using pairs of subgoals instead of random pairs. In addition to the baselines capable of solving TSP, the performance of our method 
is compared to a known approximation algorithm Christofides \cite{christofides1976worst}, to the training dataset (Teacher), and to an Oracle variant of subgoal-level MCTS where we replace the VQVAE generator with procedurally generated subgoals, where the agent is visiting one of the remaining unvisited cities.

The solution lengths found in TSP are reported in Table~\ref{tab:solutionRateTSP}. HIPS finds better solutions than the baselines in TSP. HIPS with PHS* can only slightly improve the training data, whereas HIPS-MCTS with subgoal-level rollouts can find significantly better solutions than the ones in the training dataset. It is inferior to the approximation algorithm, but the gap to the Oracle MCTS is small (around 6 \%), which shows that the subgoals generated by the VQVAE are competitive with the procedurally generated subgoals. Finally, HIPS-A* is the best-performing agent, and the gap to Christofides is around 20 \%. Note that our learned heuristic is non-admissible, and we trade off optimality for speed. kSubS can also improve on the training dataset, but it is uncompetitive against HIPS-MCTS and HIPS-A*. CQL, BC, and AdaSubS can make some progress on the task and solve some instances, but they cannot improve the solution lengths. A problem of model-free baseline methods is the inability to commit to going to a specific city, which highlights the benefits of goal-conditioned learning. Complete results including the standard errors can be found in Appendix~\ref{app:results}.

\subsection{Visualizations}

We visualize the subgoals and plans generated by our agent to gain further understanding into the agent. An example of a high-level plan found by HIPS for Sokoban is visualized in Fig.~\ref{fig:sokoban_subgoals_text}. Fig.~\ref{fig:tsp_subgoals_short} illustrates the subgoals proposed by the HIPS generative model for an intermediate state in TSP. The model suggests visiting one of the unvisited cities as the next subgoal, which is a very reasonable planning strategy.
More visualizations of the high-level trajectories discovered by our agent and the subgoals proposed by the generative model can be found in Appendix~\ref{app:hyperparameters}.

\section{Conclusion}

We present a novel method for hierarchical IL that can address difficult reasoning problems that require long-term decision-making. Our approach relies on identifying subgoals from trajectories and generating new subgoals for search-based planning on new problem instances. Our method outperforms powerful search, IL, and offline RL baselines on the benchmark tasks. 
Our experiments demonstrate that our VQVAE is a suitable generative model for subgoal-level search and using a detector to discover subgoals has benefits over subtrajectories of fixed length.

We see many promising directions for addressing the limitations of our method and improving it. Quantifying the model uncertainty could help prevent the search from exploiting the learned models. Combining discrete high-level planning with continuous low-level execution could make it possible to solve real-world tasks with robotics. Learning more abstract goals not formulated in the observation space to improve the efficiency of high-level planning and allowing the agent to ignore task-irrelevant sensory inputs to handle real-world vision tasks are also left for future work. Combining low-level and high-level searches would improve the solution rate. Applying the learned high-level models in an RL setting to improve exploration or in a curriculum learning to solve progressively harder problem instances are also promising directions for future research.

\section*{Acknowledgements}

We acknowledge the computational resources provided by the Aalto Science-IT project and CSC, Finnish IT Center for Science. The work was funded by Academy of Finland within the Flagship Programme, Finnish Center for Artificial Intelligence (FCAI).
J.~Pajarinen was partly supported by Academy of Finland (345521).




\bibliography{paper}
\bibliographystyle{icml2023}

\newpage
\appendix
\onecolumn

\section{Ethical Considerations and Societal Impact}

We do not see any immediate negative societal impacts associated with our work. We do not train our models with any sensitive or private data, and our model is not directly applicable to, for instance, real-world decision-making concerning humans. However, we cannot exclude the method being applied to something harmful that is difficult to foresee, for instance, military purposes.

\section {Infrastructure}

We trained our models on an HPC cluster using one NVIDIA GPU and multiple CPU workers per run. Most runs were performed on V100 GPUs with 32 GB of GDDR SDRAM. For some of the runs, the GPU was an A100, a K100, or a P80. We used 6 CPU workers per GPU with 10 GB of RAM per worker and each worker running on one core. By reducing the number of workers, it is possible to train and evaluate the agent on a workstation with Intel i7-8086K, 16 GB of RAM, and an NVIDIA GeForce GTX 1080 Ti GPU with 10 GB of video memory.

\section{Tried Ideas}

During the development of the algorithm, we evaluated the different components, and we found that many of the elements in the presented solution are extremely important for achieving good performance. We tried replacing the detector with a straight-through estimator but found it very difficult to train, and it did not achieve good results. We also tried Implicit Maximum Likelihood Estimation \citep[I-MLE,][]{niepert2021implicit} for trajectory segmentation, but it failed to learn meaningful representations. We found that training the detector with PPO~\citep{schulman2017proximal} instead of REINFORCE did not have a significant effect on the performance in our initial tests, so we chose REINFORCE for the conceptual simplicity. Having fixed-length subgoals performed inferior to the detector, as the ablation (see Appendix~\ref{app:k}) study shows, and training a non-goal conditioned low-level policy did not work with our algorithm either. We also tried replacing the VQVAE with an autoregressive model, but it was not computationally feasible. A GAN-based generative model was difficult to train and could not easily generate subgoals that were dissimilar enough.

We attempted not evaluating reachability during planning. However, even though the generative model has been successfully trained, the fact that we use the same number of codes regardless of the state causes the generator to create some unreachable subgoals in states where the number of realistic subgoals is smaller than the number of latent codes or some latent code corresponds to, for instance, movement in an impossible direction. Given the limited number of trajectories, we do not have enough data to train the VQVAE prior to accurately reflect this. Hence, reachability must also be addressed during planning. We tried training a reachability network to achieve this, but the performance was not good enough in practice to rely solely on it. However, using a verifier to support the reachability evaluation as in \cite{zawalski2022fast} could improve the performance of our method. 

\clearpage

\section{Impact of Value Function and VQVAE Prior}

We ran an ablation study in Sokoban to analyze the value function and the VQVAE prior. When the algorithm uses both the value function and the prior, the natural choice for the search algorithm is PHS*. When there is no value function, PHS* cannot be used, and the most similar suitable alternative to PHS* is LevinTS \citep{orseau2018single}. With a value function but without a prior, neither LevinTS nor PHS* cannot be used, so we chose GBFS as a representative search algorithm. The results for the number of puzzles solved after $N$ expansions are plotted in Table~\ref{tab:ablationValue}. The results of the ablation show that HIPS-PHS* is superior to HIPS-GBFS, indicating that learning a VQVAE prior is beneficial for solving Sokoban. The comparison between HIPS-PHS* and HIPS-LevinTS shows that value functions are important for quickly solving the puzzles, as the inferior performance after 100 node expansions shows, but the impact at 500 expansions and after that is smaller. Overall having both a value function and the VQVAE prior increases performance in Sokoban, but using one suffices for achieving a practically functional algorithm.

\begin{table*}[htpb]
\caption{The success rates (\%, higher is better) of HIPS (with PHS*), HIPS-GBFS and HIPS-LevinTS algorithms after performing $N$ node expansions in Sokoban, including the standard errors of the means of the runs.}
\label{tab:ablationValue}
\centering
\begin{tabular}{l|rrrrr}
\multicolumn{1}{c|}{} & \multicolumn{5}{c}{\textbf{Sokoban}} \\
\midrule
$N$ & 50 & 100 & 200 & 500 & 1000 \\
\midrule
HIPS-PHS* & \textbf{82.9} $\pm$ 0.9 & \textbf{88.5} $\pm$ 0.7 & \textbf{92.1} $\pm$ 0.7 & \textbf{94.7} $\pm$ 0.7 & \textbf{95.9} $\pm$ 0.5 \\
HIPS-GBFS & 74.4 $\pm$ 1.6 & 78.4 $\pm$ 1.5 & 82.9 $\pm$ 1.4 & 86.5 $\pm$ 1.4 & 89.7 $\pm$ 1.2 \\
HIPS-LevinTS & 63.8 $\pm$ 3.6 & 78.6 $\pm$ 2.3 & 87.0 $\pm$ 1.4 & 93.7 $\pm$ 0.6 & 95.6 $\pm$ 0.3 \\

\end{tabular}
\end{table*}

\section{Generalization Ability}

To analyze the generalizability of our method, we took the HIPS and HIPS-env agents trained on Sokoban levels with 4 boxes and evaluated them on Sokoban levels with 5 boxes without any fine-tuning, and the results in Table~\ref{tab:sokobanOOD} show that the performance stays good and our method has some generalization capability: 

\begin{table*}[htpb]
\caption{The success rates (\%, higher is better) of HIPS and HIPS-env after performing $N$ node expansions in Sokoban and the final success rate ($\infty$), including the standard errors of the means of the runs.}
\label{tab:sokobanOOD}
\centering
\begin{tabular}{l|rrrrrr}
$N$ & 50 & 100 & 200 & 500 & 1000 & $\infty$ \\
\midrule
HIPS, 4 boxes & 82.9 $\pm$ 0.9 & 88.5 $\pm$ 0.7 & 92.1 $\pm$ 0.7 & 94.7 $\pm$ 0.7 & 95.9 $\pm$ 0.5 & 97.5 $\pm$ 0.6 \\
HIPS, 5 boxes & 67.5 $\pm$ 2.1 & 79.4 $\pm$ 1.2 & 86.0 $\pm$ 1.3 & 91.5 $\pm$ 0.8 & 93.3 $\pm$ 0.6 & 97.1 $\pm$ 0.5 \\
\midrule
HIPS-env, 4 boxes & 82.8 $\pm$ 0.6 & 88.5 $\pm$ 0.4 & 91.5 $\pm$ 0.4 & 94.9 $\pm$ 0.3 & 96.4 $\pm$ 0.3 & 98.1 $\pm$ 0.4 \\
HIPS-env, 5 boxes & 68.6 $\pm$ 1.7 & 77.3 $\pm$ 0.9 & 83.9 $\pm$ 1.0 & 89.9 $\pm$ 0.8 & 92.8 $\pm$ 0.8 & 97.3 $\pm$ 0.7 \\
\end{tabular}
\end{table*}

\clearpage

\section{Complete Results}
\label{app:results}

\begin{table}[htb!]
\caption{The overall success rates (\%) of different algorithms including the standard errors of the means of the runs. The algorithms in the bottom part have access to the true environment dynamics, and those in the upper part do not.}
\label{tab:solutionRateComplete}
\centering
\begin{tabular}{lrrrr}
Method & Sokoban & STP & BW & TSP\\
\midrule
HIPS (ours) & 97.5 $\pm$ 0.6 & \textBF{94.7} $\pm$ 1.0 & \textBF{55.7} $\pm$ 2.4 & \textBF{100.0} $\pm$ 0.0\\
HIPS-$k$ (ours) & \textBF{99.0} $\pm$ 0.3 & \textbf{94.7} $\pm$ 1.5 & 20.1 $\pm$ 1.4 & \textBF{100.0} $\pm$ 0.0 \\
BC & 18.7 $\pm$ 0.7 & 82.5 $\pm$ 2.2 & 41.1 $\pm$ 1.6 & 28.8 $\pm$ 8.5 \\
CQL & 3.3 $\pm$ 0.4 & 11.7 $\pm$ 3.3 & 6.0 $\pm$ 1.0 & 33.6 $\pm$ 2.6 \\
DT & 36.7 $\pm$ 1.2 & 0.0 $\pm$ 0.0 & 10.0 $\pm$ 2.3 & 0.0 $\pm$ 0.0\\
RIS & 0.0 $\pm$ 0.0 & 0.0 $\pm$ 0.0 & 7.3 $\pm$ 2.2  & 0.0 $\pm$ 0.0 \\
Option-GAIL & 0.3 $\pm$ 0.3 & 0.0 $\pm$ 0.0 & 0.0 $\pm$ 0.0 & 0.0 $\pm$ 0.0 \\
IQ-Learn & 0.0 $\pm$ 0.0 & 0.0 $\pm$ 0.0 & 0.0 $\pm$ 0.0 & 0.0 $\pm$ 0.0 \\
\midrule
HIPS-env (ours) & 98.1 $\pm$ 0.4 & 94.6 $\pm$ 1.0 & 99.6 $\pm$ 0.2 & 100.0 $\pm$ 0.0 \\ %
AdaSubS & 91.4 $\pm$ 0.5 & 0.0 $\pm$ 0.0 & & 22.4 $\pm$ 2.3 \\
kSubS & 90.5 $\pm$ 1.0 & 93.3 $\pm$ 0.9 & & 87.9 $\pm$ 3.8 \\
\end{tabular}
\vspace{8 mm}
\caption{The interquartile means of the success rates (\%, higher is better) of HIPS after performing $N$ node expansions, and the final success rate ($\infty$ expansions)}
\label{tab:searchExpansionsIQM}
\centering
\begin{tabular}{l|rrrrrr}
$N$ & 50 & 100 & 200 & 500 & 1000 & $\infty$ \\
\midrule
Sokoban & 82.9 & 88.9 & 92.8 & 95.3 & 96.2 & 98.1 \\
Sliding Tile Puzzle & 79.1 & 90.6 & 92.0 & 92.7 & 93.1 & 94.6 \\
\midrule
$N$ & 20 & 50 & 100 & 500 & 1000 & $\infty$ \\
\midrule
Box-World & 56.1 & 56.2 & 56.2 & 56.2 & 56.2 & 56.2 \\
Travelling Salesman & 58.1 & 100.0 & 100.0 & 100.0 & 100.0 & 100.0 \\
\end{tabular}
\vspace{8 mm}
\caption{The success rates of different algorithms (higher is better) and the average number of steps (lower is better) needed to solve TSP. The table also includes the standard errors of the means of the runs.}
\centering
\begin{tabular}{lrr}
 Method & Success rate (\%) & Avg. steps \\ 
 \midrule
HIPS-PHS* (ours) & \textBF{100.0} $\pm$ 0.0 & 305.9 $\pm$ 5.4 \\ 
 HIPS-MCTS (ours) & \textBF{100.0} $\pm$ 0.0 & 213.0 $\pm$ 4.4\\ 
 HIPS-A* (ours) & \textBF{100.0} $\pm$ 0.0 & \textBF{168.2} $\pm$ 3.6 \\
 kSubS & 87.9 $\pm$ 3.8 & 268.2 $\pm$ 12.0 \\
 CQL & 33.6 $\pm$ 2.6 & 336.9 $\pm$ 4.1 \\ 
 BC & 28.8 $\pm$ 8.5 & 339.3 $\pm$ 3.9 \\ 
 AdaSubS & 22.4 $\pm$ 2.3 & 338.9 $\pm$ 18.6 \\
 \midrule
 Teacher & 100.0 $\pm$ 0.0 & 336.5 $\pm$ 0.4 \\
 Oracle MCTS & 100.0 $\pm$ 0.0 & 199.9 $\pm$ 0.7 \\
 Christofides & 100.0 $\pm$ 0.0 & 139.0 $\pm$ 0.1 \\
\end{tabular}
\label{tab:solutionRateTSPComplete}
\end{table}

\begin{table*}[htpb]
\caption{The mean success rates (\%, higher is better) of different search algorithms after performing $N$ node expansions, including the standard errors of the means of the runs.}
\label{tab:searchExpansionsComplete}
\centering
\begin{tabular}{l|rrrrr}
\multicolumn{1}{c|}{} & \multicolumn{5}{c}{\textbf{Sokoban}} \\
\midrule
$N$ & 50 & 100 & 200 & 500 & 1000 \\
\midrule
HIPS (ours) & \textbf{82.9} $\pm$ 0.9 &\textbf{88.5} $\pm$ 0.7 & \textbf{92.1} $\pm$ 0.7 & 94.7 $\pm$ 0.7 & 95.9 $\pm$ 0.5 \\
HIPS-$k$ (ours) & 63.1 $\pm$ 1.7 & 77.7 $\pm$ 1.0 & 85.8 $\pm$ 0.6 & 90.9 $\pm$ 0.6 & 94.3 $\pm$ 0.0 \\
\midrule
HIPS-env (ours) & 82.8 $\pm$ 0.6 & \textBF{88.5} $\pm$ 0.4 & 91.5 $\pm$ 0.4 & \textBF{94.9} $\pm$ 0.3 & \textBF{96.4} $\pm$ 0.3 \\
HIPS-env-$k$ (ours) & 64.3 $\pm$ 1.2 & 76.9 $\pm$ 0.8 & 86.3 $\pm$ 1.1 & 91.3 $\pm$ 1.5 & 94.0 $\pm$ 1.6 \\
AdaSubS & 76.4 $\pm$ 0.5 & 82.2 $\pm$ 0.5 & 85.7 $\pm$ 0.6 & 88.8 $\pm$ 0.4 & 90.2 $\pm$ 0.5 \\ 
kSubS & 69.1 $\pm$ 2.2 & 73.1 $\pm$ 2.2 & 76.3 $\pm$ 1.9 & 79.9 $\pm$ 2.2 & 82.8 $\pm$ 1.3 \\
PHS* & 0.3 $\pm$ 0.0 & 1.8 $\pm$ 0.1 & 14.9 $\pm$ 0.3 & 76.1 $\pm$ 0.5 & 91.1 $\pm$ 0.3 \\
\end{tabular}
\vspace{5 mm}
\begin{tabular}{l|rrrrr}
\multicolumn{1}{c|}{} & \multicolumn{5}{c}{\textbf{Sliding Tile Puzzle}} \\
\midrule
$N$ & 50 & 100 & 200 & 500 & 1000 \\
\midrule
HIPS (ours) & \textbf{78.7} $\pm$ 2.5 & 89.8 $\pm$ 1.9 & 91.1 $\pm$ 1.7 & 91.8 $\pm$ 1.7 & 92.2 $\pm$ 1.7 \\
HIPS-$k$ (ours) & 10.3 $\pm$ 1.2 & 68.0 $\pm$ 3.9 & 76.0 $\pm$ 3.8 & 79.1 $\pm$ 3.5 & 84.4 $\pm$ 1.6 \\
\midrule
HIPS-env (ours) & 78.6 $\pm$ 2.6 & \textBF{90.2} $\pm$ 1.8 & \textbf{91.4} $\pm$ 1.6 & \textBF{92.1} $\pm$ 1.7 & 92.6 $\pm$ 1.6 \\
HIPS-env-$k$ (ours) & 10.5 $\pm$ 1.8 & 69.7 $\pm$ 2.2 & 76.6 $\pm$ 3.0 & 80.9 $\pm$ 2.9 & 87.2 $\pm$ 2.3 \\
AdaSubS & 0.0 $\pm$ 0.0 & 0.0 $\pm$ 0.0 & 0.0 $\pm$ 0.0 & 0.0 $\pm$ 0.0 & 0.0 $\pm$ 0.0 \\ 
kSubS & 0.7 $\pm$ 0.2 & 79.9 $\pm$ 3.1 & 89.8 $\pm$ 1.5 & 91.7 $\pm$ 1.3 & \textbf{92.7} $\pm$ 1.1 \\
PHS* & 0.0 $\pm$ 0.0 & 0.0 $\pm$ 0.0 & 0.0 $\pm$ 0.0 & 0.0 $\pm$ 0.0 & 0.0 $\pm$ 0.0 \\
\end{tabular}
\vspace{5 mm}
\begin{tabular}{l|rrrrr}
\multicolumn{1}{c|}{} & \multicolumn{5}{c}{\textbf{Box-World}} \\
\midrule
$N$ & 20 & 50 & 100 & 500 & 1000 \\
\midrule
HIPS (ours) & 55.6 $\pm$ 2.3 & 55.7 $\pm$ 2.3 & 55.7 $\pm$ 2.3 & 55.7 $\pm$ 2.3 & 55.7 $\pm$ 2.3 \\
HIPS-$k$ (ours) & 20.1 $\pm$ 1.3 & 20.1 $\pm$ 1.3 & 20.1 $\pm$ 1.3 & 20.1 $\pm$ 1.3 & 20.1 $\pm$ 1.3 \\
\midrule
HIPS-env (ours) & \textBF{99.3} $\pm$ 0.2 & \textBF{99.6} $\pm$ 0.2 & \textBF{99.6} $\pm$ 0.2 & \textBF{99.6} $\pm$ 0.2 & \textBF{99.6} $\pm$ 0.2 \\
HIPS-env-$k$ (ours) & 89.1 $\pm$ 1.7 & 98.9 $\pm$ 0.1 & 99.1 $\pm$ 0.2 & 99.1 $\pm$ 0.2 & 99.1 $\pm$ 0.2 \\
PHS* & 0.6 $\pm$ 0.2 & 5.4 $\pm$ 0.5 & 25.8 $\pm$ 1.0 & 93.8 $\pm$ 0.7 & 94.0 $\pm$ 0.7 \\
\end{tabular}
\vspace{5mm}
\begin{tabular}{l|rrrrr}
\multicolumn{1}{c|}{} & \multicolumn{5}{c}{\textbf{Travelling Salesman}} \\
\midrule
$N$ & 20 & 50 & 100 & 500 & 1000 \\
\midrule
HIPS (ours) & 52.9 $\pm$ 13.3 & \textbf{99.9} $\pm$ 0.1 & \textbf{100.0} $\pm$ 0.0 & \textbf{100.0} $\pm$ 0.0 & \textbf{100.0} $\pm$ 0.0 \\
HIPS-$k$ (ours) & 0.0 $\pm$ 0.0 & 12.5 $\pm$ 1.4 & 77.3 $\pm$ 1.8 & \textbf{100.0} $\pm$ 0.0 & \textbf{100.0} $\pm$ 0.0\\
\midrule
HIPS-env (ours) & \textbf{54.3} $\pm$ 13.7 & \textbf{99.9} $\pm$ 0.1 & \textbf{100.0} $\pm$ 0.0 & \textBF{100.0} $\pm$ 0.0 & \textBF{100.0} $\pm$ 0.0 \\
HIPS-env-$k$ (ours) & 0.0 $\pm$ 0.0 & 12.9 $\pm$ 1.7 & 77.2 $\pm$ 1.6 & \textBF{100.0} $\pm$ 0.0 & \textBF{100.0} $\pm$ 0.0 \\
AdaSubS & 0.0 $\pm$ 0.0 & 0.0 $\pm$ 0.0 & 1.2 $\pm$ 0.6 & 12.2 $\pm$ 1.6 & 20.8 $\pm$ 1.2 \\ 
kSubS & 0.0 $\pm$ 0.0 & 1.5 $\pm$ 0.6 & 40.4 $\pm$ 11.1 & 80.9 $\pm$ 6.8 & 85.3 $\pm$ 5.3 \\
PHS* & 0.0 $\pm$ 0.0 & 0.0 $\pm$ 0.0 & 0.0 $\pm$ 0.0 & 0.0 $\pm$ 0.0 & 0.0 $\pm$ 0.0\\
\end{tabular}
\end{table*}

\clearpage

\section{Ablation: HIPS-$k$}
\label{app:k}

\begin{table*}[htpb]
\caption{The success rates (\%, higher is better) of different variants of HIPS in STP after performing $N$ node expansions}
\centering
\begin{tabular}{l|rrrr}
$N$ & 100 & 500 & 1000 & $\infty$\\
\midrule
HIPS-PHS* & 55.8 $\pm$ 4.5 & \textbf{94.6} $\pm$ 0.8 & \textbf{94.8} $\pm$ 0.9 & \textbf{94.8} $\pm$ 0.9 \\
HIPS-GBFS & \textbf{89.8} $\pm$ 1.9 & 91.8 $\pm$ 1.7 & 92.2 $\pm$ 1.7 & 94.7 $\pm$ 1.0\\
\midrule
HIPS-GBFS-3 & N/A & N/A & N/A & N/A \\
HIPS-GBFS-5 & 68.1 $\pm$ 3.9 & 79.1 $\pm$ 3.2 & 84.4 $\pm$ 1.4 & 94.7 $\pm$ 1.4 \\
HIPS-GBFS-7 & 84.3 $\pm$ 0.4 & 85.6 $\pm$ 0.5 & 86.1 $\pm$ 0.6 & 86.4 $\pm$ 0.5 \\
HIPS-GBFS-9 & 76.4 $\pm$ 0.8 & 76.4 $\pm$ 0.8 & 76.4 $\pm$ 0.8 & 76.4 $\pm$ 0.8 \\
\end{tabular}
\label{tab:stp_k}
\end{table*}

When we replace the detector $d_\xi$ with subgoals sampled at fixed intervals, we re-train the low-level policy $\pi_\theta$ to achieve these subgoals. The discrete VQVAE, including the prior, are also re-trained using the new pairs of consecutive subgoals. The distance between the subgoals can, in this case, be controlled by a hyperparameter $k$. In Sokoban, STP, and Box-World, we used ten as the subgoal horizon and five as $k$ (see Table~\ref{tab:hyperparameters}). In TSP, selecting a segment length half of the subgoal horizon proved to be too much, so we let $k$ be equal to four, the default segment length used in kSubS \cite{czechowski2021subgoal}. 

\begin{figure}[b!]
\centering
\begin{minipage}{0.35\textwidth}
\centering
\includegraphics[width=\textwidth,trim={5mm 8mm 10mm 10mm},clip]{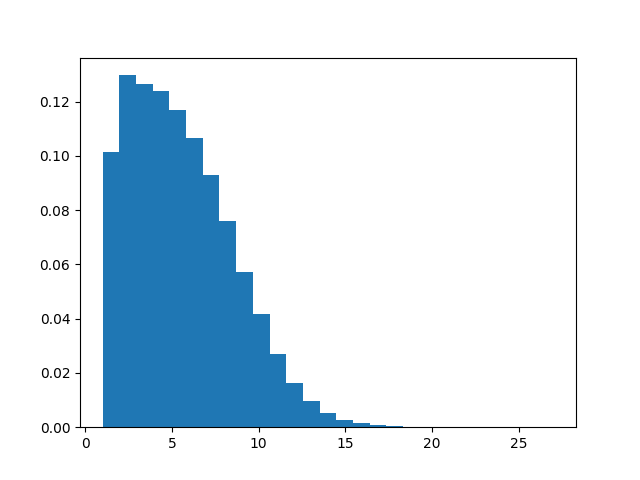}
\\
Sokoban
\end{minipage}
\begin{minipage}{0.35\textwidth}
\centering
\includegraphics[width=\textwidth,trim={5mm 8mm 10mm 10mm},clip]{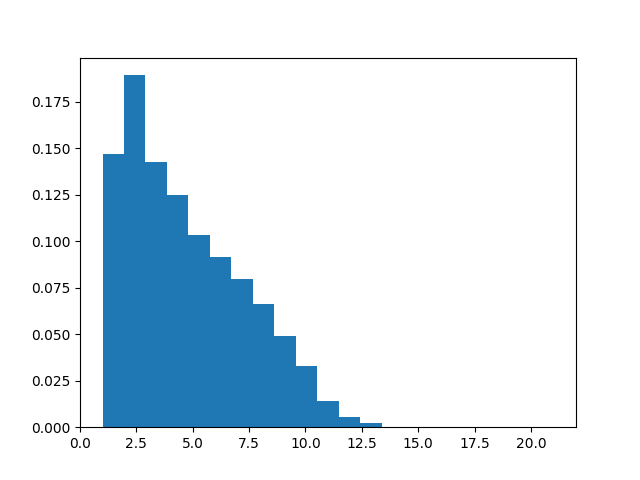}
\\
STP
\end{minipage}
\\
\begin{minipage}{0.35\textwidth}\centering
\includegraphics[width=\textwidth,trim={5mm 8mm 10mm 10mm},clip]{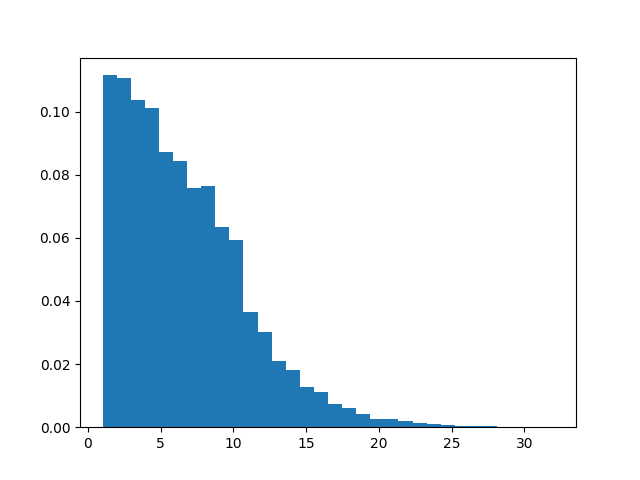}
\\
Box-World
\end{minipage}
\begin{minipage}{0.35\textwidth}
\centering
\includegraphics[width=\textwidth,trim={5mm 8mm 10mm 10mm},clip]{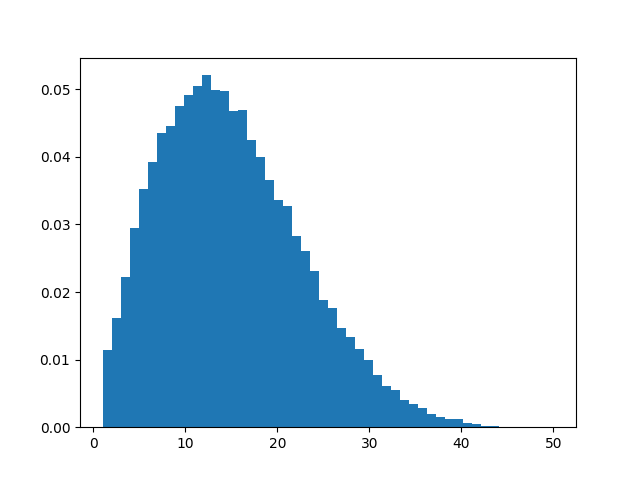}
\\
TSP
\end{minipage}
\caption{Lengths of the subtrajectories to reach subgoals proposed by VQVAE when it has been trained using the detector $d_\xi$.}
\label{fig:k}
\end{figure}

We performed a small ablation study in STP to analyze the impact of the value of $k$. The results are shown in Table~\ref{tab:stp_k}. We see that given a large research budget, PHS* is slightly superior to GBFS, but GBFS outperforms PHS* given a small search budget. HIPS-GBFS-3 doesn't converge because the value function is noisy. Using a larger $k$ allows the value function to "leap over" the noise, as observed by \citet{czechowski2021subgoal}. We see that using a larger $k$ improves the percentage of puzzles solved after a smaller number of expansions, but hurts the overall solution rate, as the search space is explored less systematically. However, using a $k$ too large is also harmful as training the generative model becomes difficult. Furthermore, no value of $k$ was able to outperform HIPS-GBFS, which highlights the benefits of our method that can propose subgoals at different distances adaptively in all environments (see Figure~\ref{fig:k}).

\clearpage

\section{Search Methods}
\label{app:search}

\paragraph{Greedy Best-First Search (GBFS)} is a priority queue -based search algorithm, where the evaluation function has been defined as
\begin{align*}
    \varphi(n) = h(n),
\end{align*}
where $h(n)$ is a heuristic that predicts the distance to the goal. The node that is predicted to be the closest to the goal is expanded.

\paragraph{Policy-Guided Heuristic Search (PHS)} is a policy-guided search algorithm \citep{orseau2021policy}
which uses a priority queue with the evaluation function
\begin{align*}
    \varphi(n) = \eta(n) g(n) / \pi(n),
\end{align*}
where $g(n)$ is the path cost from the root to node $n$, $\pi(n)$ is the node policy (the probability of selecting node $n$) and
$\eta(n)$ is a heuristic factor whose purpose is to estimate
the cost to the nearest descendant solution node. We use a variant of PHS, PHS* where the heuristic factor has been defined as
\begin{equation}
\eta_h(n) = \frac{1 + h(n)/g(n)}{\pi(n)^{h(n) / g(n)}}
\,.
\label{eq:phs_star}
\end{equation}

\paragraph{A*} is a heuristic-based search algorithm that tries to find the shortest path to the goal \cite{hart1968formal}. It is based on a priority queue with the evaluation function
\begin{align*}
    \varphi(n) = g(n) + h(n),
\end{align*}
where $g(n)$ is the distance from the root to node $n$ and $h(n)$ is a heuristic that predicts the distance from the node $n$ to the goal. If the heuristic $h(n)$ never overestimates the true distance to the goal, the heuristic is said to be admissible, and A* is guaranteed to find the shortest path.

\paragraph{Monte Carlo Tree Search (MCTS)} is a tree-based search method based on expanding a search tree by performing Monte Carlo evaluations.
The selection of nodes for expansion is biased towards promising nodes to enable MCTS to focus on the relevant parts of the search tree. The most commonly used method is UCT, where nodes with higher rewards and lower visitation frequency get the highest priority \cite{ozair2021vector, kocsis2006bandit}.

Listing~\ref{alg:search} contains the pseudo-code describing our high-level search with a priority queue. Subroutine \texttt{phs\_cost} calculates the value of the heuristic factor $\eta(n)$ for PHS* as in \eqref{eq:phs_star}. Subroutine \texttt{extract\_plan} collects the subgoals on the path to the leaf node (terminal state) from the root (initial state). Subroutine \texttt{get\_distances} takes as input the current state and the proposed children and tries to reach them using the subgoal-conditioned low-level policy $\pi_\theta$. The distances between the state and the children are recovered simultaneously.

\begin{listing}[tb]
\caption{PyTorch pseudocode for the high-level search}
\label{alg:search}
\begin{lstlisting}[language=Python]
def get_priority(node, alg):
    if alg == 'phs_star':
        return phs_cost(node.log_p, node.value, node.cum_dist)
    elif alg == 'gbfs':
        return node.value
    elif alg == 'a_star':
        return node.value + node.cum_dist


def init_node(node, alg, vqvae, policy, value_func, dynamics):
    node.value = value_func(node.state)
    node.child_states = vqvae.generate(node.state)
    node.distances_to_children = get_distances(
        node.state,
        node.child_states,
        policy,
        dynamics)
    node.filter_unreachable_children()   # Uses the distances computed
    node.children_log_probs = vqvae.prior(node.state)
    node.priority = get_priority(node, alg)


def search(state, alg, vqvae, policy, value_func, dynamics):
    n_nodes = 0
    queue = PriorityQueue()   # Create empty priority queue
    expanded = Set()   # Create empty set
    node = Node(
        state,
        parent=None,
        cum_dist=0,
        log_p=0,
    )
    init_node(node, alg, vqvae, policy, value_func, dynamics)
    queue.insert(node)

    while len(queue) > 0:
        node = queue.pop()
        expanded.add(node.state)
        n_nodes += 1
        for c_state, c_dist, c_log_p in zip(node.child_states,
                                            node.distances_to_children,
                                            node.children_log_probs):
            if c_state in expanded:
                continue
            new_node = Node(
                c_state,
                parent=node,
                cum_dist=node.cum_dist + c_dist,
                log_p=node.log_p + c_log_p,
            )
            if is_terminal(c_state):
                return extract_plan(new_node), n_nodes   # Success
            init_node(new_node, alg, vqvae, policy, value_func, dynamics)
            queue.insert(new_node)
    return None, n_nodes   # Search failed, queue empty
\end{lstlisting}
\end{listing}

\clearpage

\section{VQVAE Training}
\label{app:vqvae}

\begin{algorithm}
\caption{Training VQVAE for Subgoal Generation}
\label{alg:vqvae}
\textbf{Input}: A dataset of trajectories $\mathcal{D}$, a dataset of subgoal pairs~$\mathcal{D}'$, untrained encoder $f_\phi$, decoder $g_\psi$, codebook $\{e_k\}$ \\
\textbf{Parameters}: The codebook $\{e_k\}$ and the parameters of the encoder, $\phi$, and the decoder, $\psi$ \\
\textbf{Output}: Trained encoder $f_\phi$, decoder $g_\psi$, codebook~$\{e_k\}$. 
\begin{algorithmic}[1] 
\WHILE{$f_\phi$ and $g_\psi$ not converged}
\STATE Sample a trajectory $\tau$ from $\mathcal{D}$
\STATE For each state $s_i \in \tau$, uniformly sample a pair $s_j$ from $(s_{i+1}, \dots, s_{H})$
\STATE Reconstruct $s_j$ without discretization: $\hat{s}_j = g_\psi(f_\phi(s_j, s_i), s_i)$
\STATE Compute reconstruction loss $\mathcal{L}_\text{rec}(\hat{s}_j, s_j)$ 
\STATE Update $\phi$, $\psi$ to minimize reconstruction loss
\ENDWHILE
\STATE Sample subsequent subgoal pairs $s_{g_{j}}$, $s_{g_{j-1}}$ from $\mathcal{D}'$ and encode them with the encoder: $z_j~=~f_\phi(s_{g_{j}}, s_{g_{j-1}})$ 
\STATE Initialize $\{e_k\}$ as the clusters centers obtained by running KMeans++ on the encodings $\{z_j\}$
\WHILE{$\{e_k\}$, $f_\phi$ and $g_\psi$ not converged}
    \STATE Sample a batch of subsequent subgoal pairs $s_{g_{j}}$, $s_{g_{j-1}}$ from $\mathcal{D}'$
    \STATE Reconstruct subgoals with VQVAE
    \begin{align*}
      z_j &= f_\phi(s_{g_{j}}, s_{g_{j-1}}) \\
      k_j &= \argmin_m \| z_j - e_m\|_2 \\
      \hat{s}_{g_{j}} &= g_\psi(e_{k_j}, s_{g_{j-1}})
    \end{align*}
    \STATE Compute the loss in \eqref{eq:vqvae_loss}
    \STATE Update $\phi$, $\psi$, $\{e_k\}$ to minimize the loss computed in Step 13.
\ENDWHILE
\STATE \textbf{return} $f_\phi$, $g_\psi$, $\{e_k\}$
\end{algorithmic}
\end{algorithm}

\section{Hyperparameters and Visualizations}
\label{app:hyperparameters}

\begin{table}[htbp]
\caption{General hyperparameters of our method.}
\centering
\begin{tabular}{lc}
Parameter & Value \\
\midrule
Learning rate for dynamics & $2 \cdot 10^{-4}$ \\
Learning rate for $\pi$, $d$, $V$ & $1 \cdot 10^{-3}$ \\
Learning rate for VQVAE & $2 \cdot 10^{-4}$ \\
Discount rate for REINFORCE & 0.99 \\
\end{tabular}
\label{tab:genhyperparameters}
\end{table}

\begin{table}[b!]
\caption{Environment-specific hyperparameters of our method.}
\centering
\begin{tabular}{llcccc}
Parameter & Explanation & Sokoban & STP & Box-World & TSP \\
\midrule
$\alpha$ & Subgoal penalty & 0.1 & 0.1 & 0.1 & 0.05 \\

$\beta$ & Beta for VQVAE & 0.1 & 0.1 & 0.1 & 0 \\
$c$ & Exploration constant for MCTS & -- & -- & -- & 0.1 \\
$D$ & Codebook dimensionality & 128 & 128 & 128 & 64 \\
$H$ & Subgoal horizon & 10 & 10 & 10 & 50 \\
$K$ & VQVAE codebook size & 64 & 64 & 64 & 32 \\
$k$ & Segment length w/o REINFORCE & 5 & 5 & 5 & 4 \\
$(N, D)$ & DRC size & -- & -- & (3, 3) & -- \\
\end{tabular}
\label{tab:hyperparameters}
\end{table}

\clearpage

\begin{figure*}[tp]
\centering

\includegraphics[width=.95\textwidth,trim={0mm 0mm 0mm 0mm},clip]{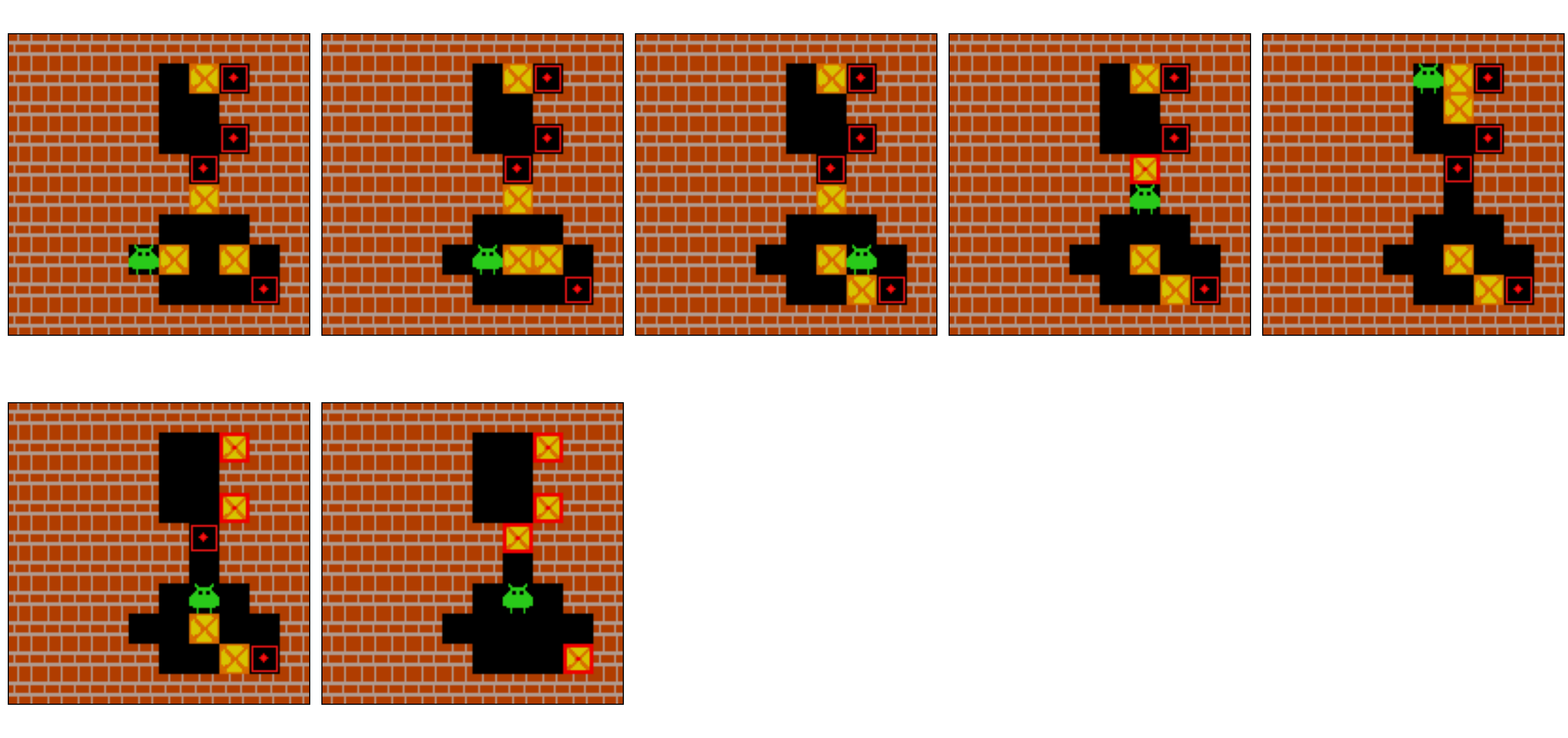}

(a) A subgoal-level plan.
\vspace{5mm}

\includegraphics[width=.95\textwidth,trim={0mm 0mm 0mm 0mm},clip]{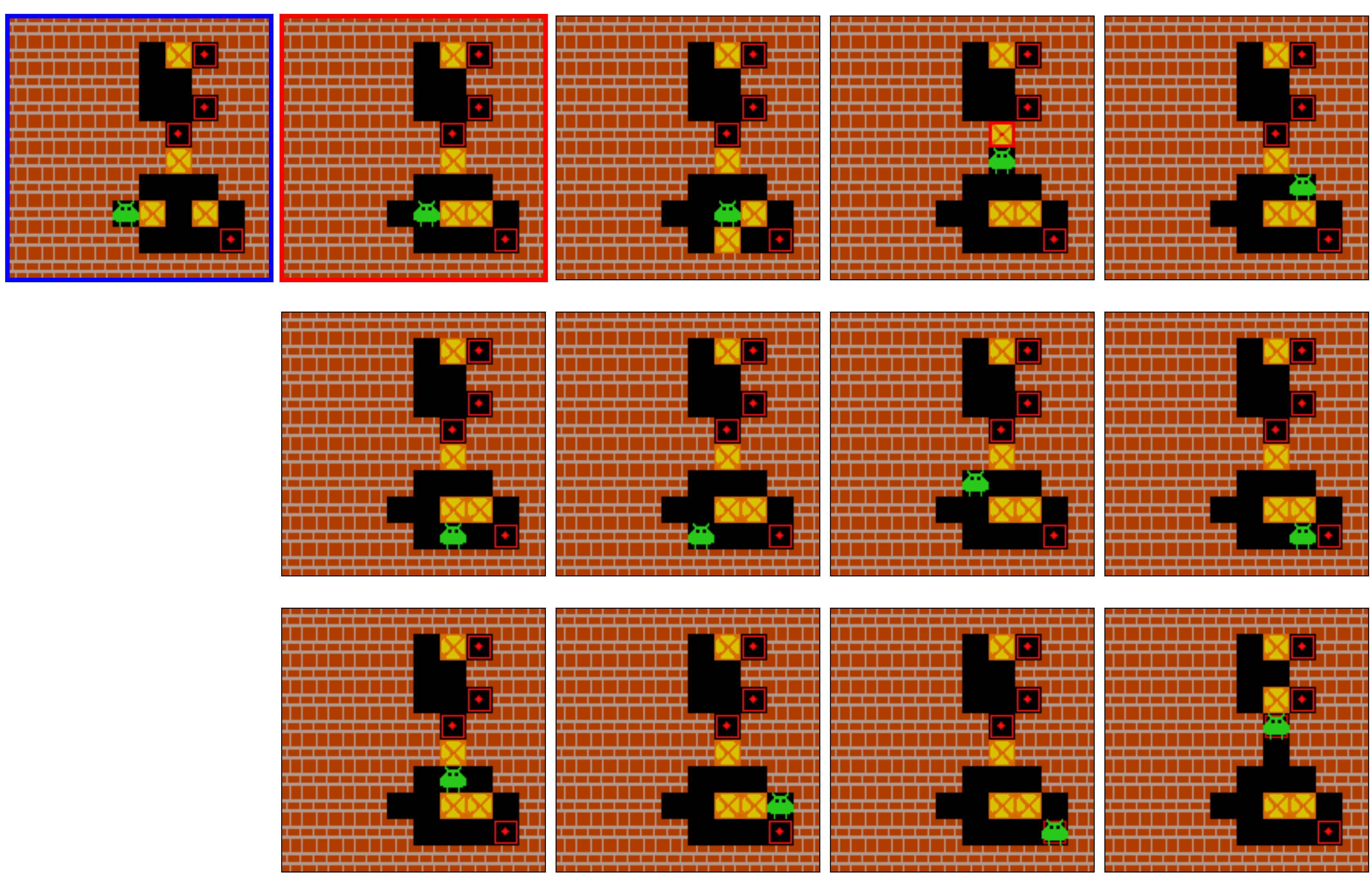}

(b) Examples of generated subgoals.

\caption{Visualization of the solution found by HIPS for a Sokoban problem.
(a):~A subgoal-level plan found by HIPS.
(b):~Subgoals proposed for an intermediate state (marked with blue boundaries). The subgoals have been sorted according to the prior probabilities. The subgoal selected for the final plan is marked with red boundaries.
}
\label{fig:sokoban_subgoals_app}
\end{figure*}

\begin{figure*}[tp]
\centering

\includegraphics[width=.95\textwidth,trim={0mm 0mm 0mm 0mm},clip]{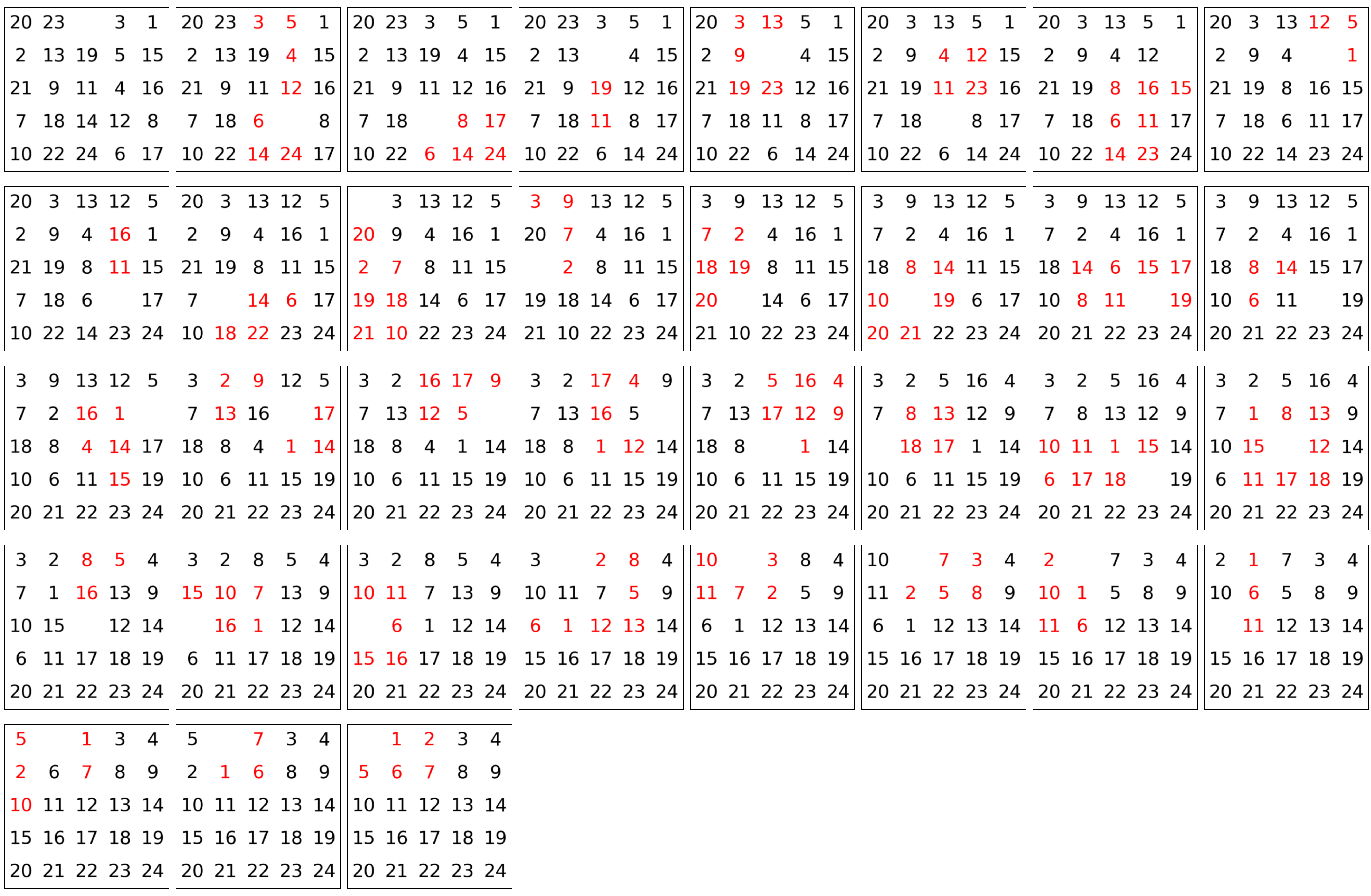}

(a) A subgoal-level plan.
\vspace{5mm}

\begin{minipage}{.95\textwidth}
    
\includegraphics[width=.985\textwidth,trim={0mm 8mm 0mm 23mm},clip]{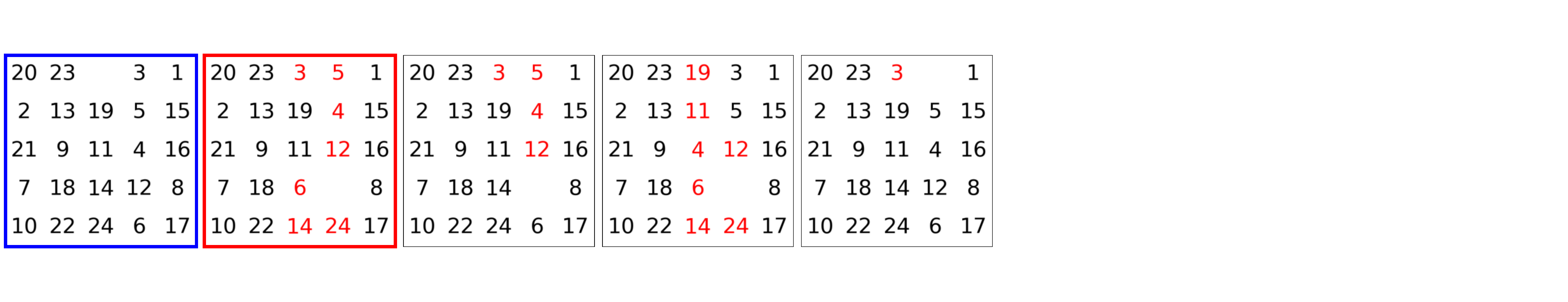}
\\[-2mm]
\includegraphics[width=.985\textwidth,trim={0mm 8mm 0mm 23mm},clip]{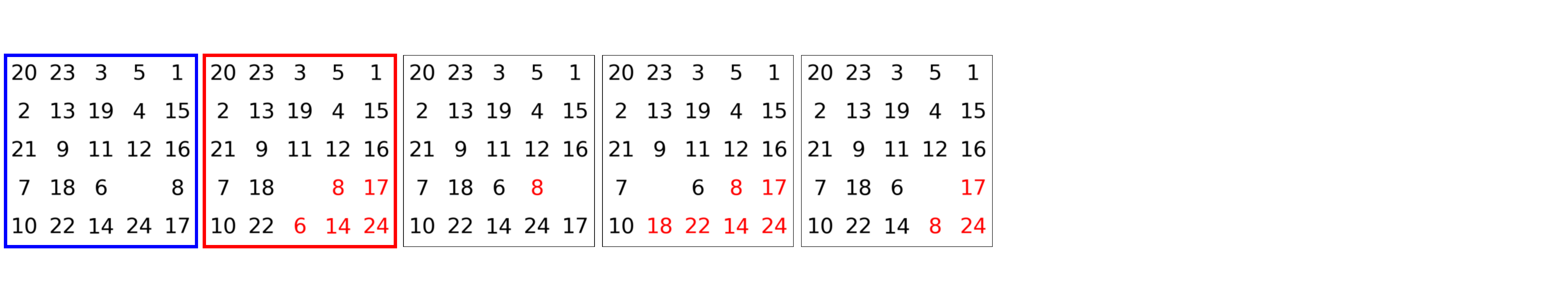}
\\[-2mm]
\includegraphics[width=\textwidth,trim={0mm 112mm 0mm 10mm},clip]{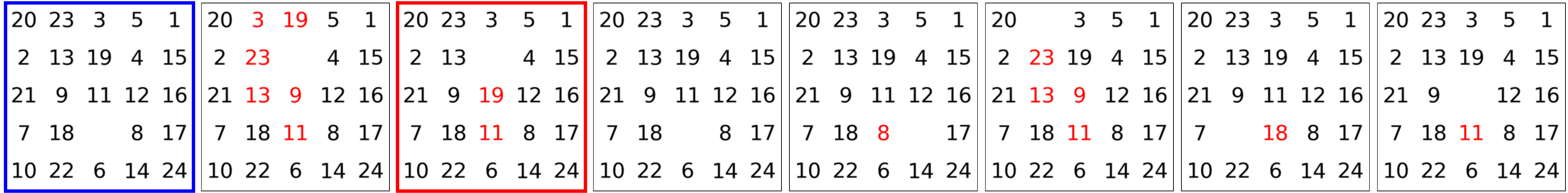}
\vspace{1mm}
\end{minipage}

(b) Examples of generated subgoals.

\caption{Visualization of the solution found by HIPS for an STP problem.
(a):~A subgoal-level plan found by HIPS.
(b):~Subgoals proposed for intermediate states (marked with blue boundaries). The subgoals have been sorted according to the prior probabilities. The subgoal selected for the final plan is marked with red boundaries.
Red color is used to highlight the tiles which are different from the reference state (the previous state in (a) and the current state in (b)).
}
\label{fig:stp_subgoals}
\end{figure*}

\begin{figure*}[tp]
\centering

\includegraphics[width=.95\textwidth,trim={0mm 21mm 0mm 21mm},clip]{Figures/bw_traj_1_bennnn.pdf}

(a) A subgoal-level plan.

\vspace{5mm}

\includegraphics[width=.95\textwidth]{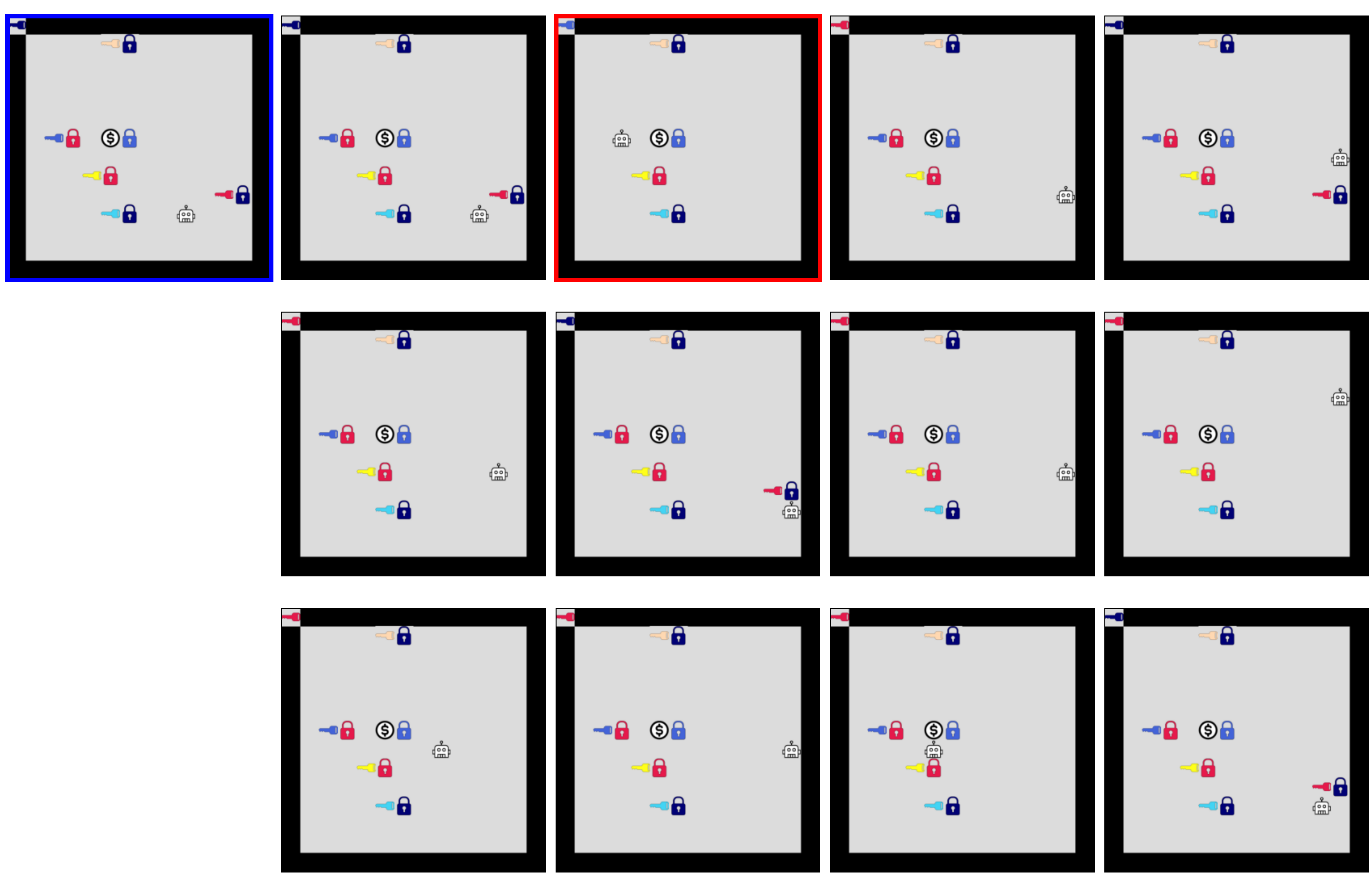}

(b) Examples of generated subgoals.

\caption{Visualization of the solution found by HIPS for a BW problem.
(a):~A subgoal-level plan found by HIPS.
(b):~Subgoals proposed for an intermediate state (marked with blue boundaries). The subgoals have been sorted according to the prior probabilities. The subgoal selected for the final plan is marked with red boundaries.
}
\label{fig:bw_vqvae}
\end{figure*}

\begin{figure*}[thpb]
\centering

\includegraphics[width=\textwidth,trim={0mm 0mm 0mm 0mm},clip]{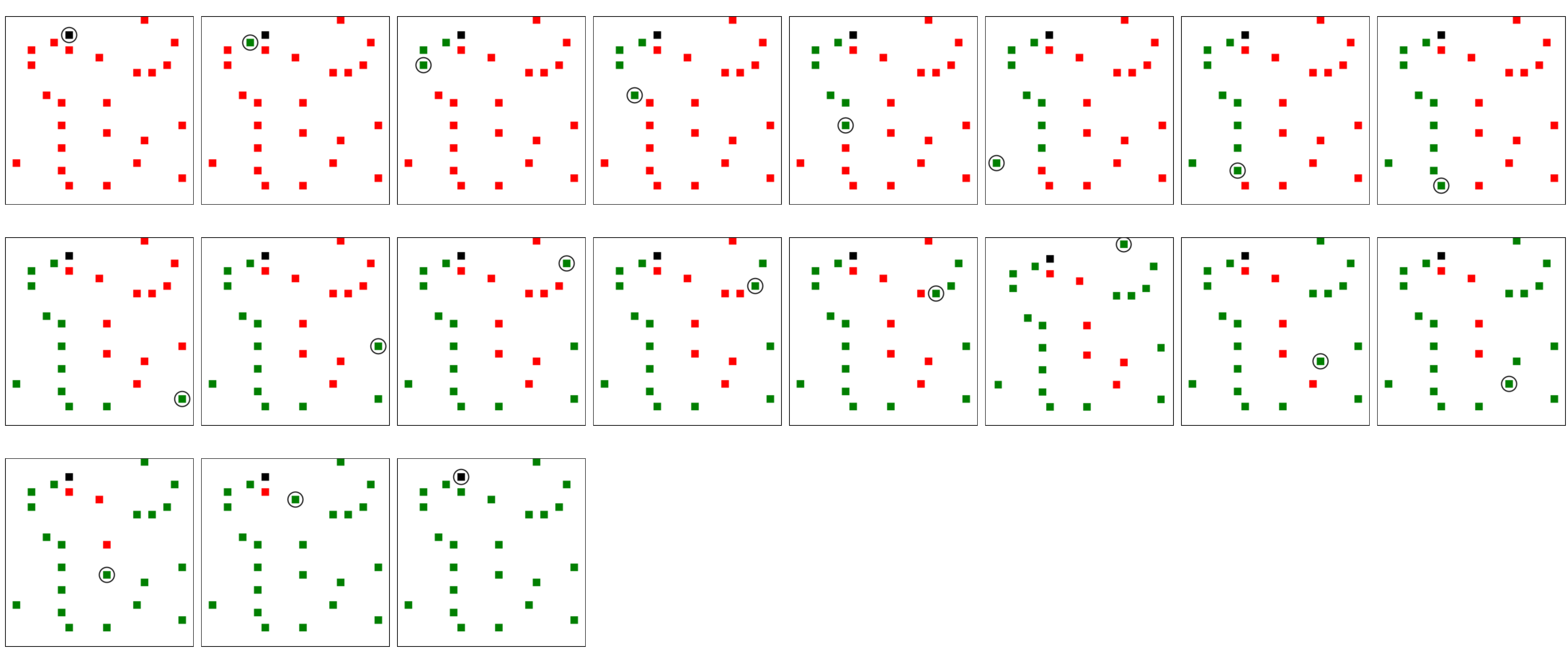}

(a) A subgoal-level plan.

\vspace{5mm}

\includegraphics[width=\textwidth,trim={0mm 0mm 0mm 0mm},clip]{Figures/tsp_traj_1_FFvVbj_n15.pdf}

(b) Examples of generated subgoals.

\caption{Visualization of the solution found by HIPS for a TSP instance.
(a):~A subgoal-level plan found by HIPS.
(b):~Subgoals proposed for an intermediate state (marked with blue boundaries).
}
\label{fig:tsp_subgoals}
\end{figure*}


\end{document}